\newcommand\SEKIusersusepackages{\usepackage{latexsym,amssymb}}
\newcommand\SEKIREVIEWSTATUS{1}
\newcommand\SEKIREVIEWER{%
\bussottiname
\\Via Paolo Lilla 66, 57122 Livorno, Italy
\\E-mail: {\tt paolo.bussotti@alice.it}
\\
}
\newcommand\SEKIDOCUMENTCLASS{0}
\newcommand\SEKIYEAR{2006}
\newcommand\SEKINUMBEROFYEAR{02}
\newcommand\SEKITYPE{1}
\newcommand\SEKIPUBLISHEDTYPE{0}
\title{A Self-Contained and\\Easily Accessible Discussion\\of the Method of\emph\DescenteInfinie\ and\\\fermat's Only Explicitly Known Proof
\\by\emph\DescenteInfinie}
\author{\wirthname
\\{\footnotesize\Institute}
\\{\small\tt\email}}

\newcommand\daspaper{paper}

\input seki-deckblatt-4
\newlength{\sectionintocsep} 
\newlength{\contentsandreferencesheadroom} 
\newlength{\contentsandreferencesfootroom} 
\catcode`\@=11

\def\tableofcontents{\ignorespaces
\section*{\contentsname\@mkboth
{\uppercase{\contentsname}}{\uppercase{\contentsname}}}%
\vspace*\contentsandreferencesfootroom
\@starttoc{toc}}
\def\contentsname{Contents} 

\catcode`\@=12

\input headerhot

\newcommand\Proofof{Proof of}
\renewenvironment{proof}[1]{\yestop\begin{sloppypar}\noindent
{\bf\Proofof\ {#1}}\\}{\end{sloppypar}}
\renewenvironment{proofqed}[1]
{\begin{sloppypar}\def\fooqed{#1}\noindent{\bf\Proofof\ \fooqed}}
{\QEDbf\fooqed\end{sloppypar}}
\renewenvironment{proofparsep}[1]{\parindent=0pt\begin
{sloppypar}\noindent{\bf\Proofof\ {#1}}\nopagebreak\par}
{\end{sloppypar}}
\renewenvironment{proofparsepqed}[1]{\parindent=0pt\begin
{sloppypar}\def\fooqed{#1}\noindent{\bf\Proofof\ \fooqed}\nopagebreak\par}
{\nopagebreak\QEDbf\fooqed\end{sloppypar}}
\renewenvironment{proofshort}[1]{\begin{sloppypar}\noindent
{\bf\Proofof\ {#1}:}}{\end{sloppypar}}

{\end{enumerate}\renewcommand{\theenumi}{\arabic{enumi}}}

\mathcommand\myfootnotemark[1]{^{#1}}

\newcommand\repname{{\rm set}}
\mathapplycommand\rep\repname
\mathcommand\repr[1]{{\repname[{#1}]}}
\mathcommand\msa{\langle}
\mathcommand\mse{\rangle}
\mathcommand\msu{\,\sqcup\,}
\mathcommand\msin{{\rm\;in\;}}
\mathcommand\mssetminus{\setminus\!\!\setminus}
\mathcommand\tightmssubseteq{\sqsubseteq}
\mathcommand\mssubseteq{\ \tightmssubseteq\ }
\mathcommand\approxapprox{\approx\:\!\!\approx}
\mathcommand\quasilquasil{\,\lesssim\!\lesssim\,}
\mathcommand\quasibquasib{\,\gtrsim\!\gtrsim\,}
\mathcommand\fmul[1]{{\rm FMul}(#1)}
\mathcommand\smul[1]{{\rm SMul}(#1)}
\mathcommand\multisetwith [2]{\msa\ {#1}\ |\ {#2}\ \mse}
\mathcommand\multisetwithq[3]{\msa\ {#2}\ |_{#1}\ {#3}\ \mse}
\newcommand\superseteq     {\supseteq}

\mathcommand\quasilhd
{\mathop{\mbox{\raisebox{0.31ex}{\math\lhd}\hspace{-0.75em}\raisebox
{-0.6ex}{\math\sim}}}}
\newcommand\quasirhd{\mbox{\raisebox{0.31ex}{$\rhd$}\hspace{-0.75em}\raisebox{-0.6ex}{$\sim$}}}

\mathcommand\rhdrhd{\rhd$\hspace{-0.35em}$\rhd}
\mathcommand\lhdlhd{\lhd$\hspace{-0.21em}$\lhd}

\mathcommand\quasilhdquasilhd{\quasilhd$\hspace{-0.13em}$\quasilhd}

\newcommand\hiddensubSS{_{_{\rm SS}}}
\mathcommand\antisubsum     {\rhd\hiddensubSS}
\mathcommand\notantisubsum  {\ntriangleright\hiddensubSS}
\mathcommand\subsum         {\lhd\hiddensubSS}
\mathcommand\notsubsum      {\ntriangleleft\hiddensubSS}
\mathcommand\antisubsumeq   {\trianglerighteq\hiddensubSS}
\mathcommand\subsumeq       {\trianglelefteq\hiddensubSS}
\mathcommand\quasisubsum    {\,\quasilhd\raisebox{0.1ex}{$\hiddensubSS$}}
\mathcommand\antiquasisubsum{\,\quasirhd\raisebox{0.1ex}{$\hiddensubSS$}}
\mathcommand\quasiquasisubsum{\quasisubsum\!\!\quasisubsum}
\mathcommand\antiquasiquasisubsum{\antiquasisubsum\!\!\!\antiquasisubsum}

\newcommand\hiddensubH{_{_{\rm H}}}
\newcommand\hiddensubCONS{_{_\CONS}}
\mathcommand\hql   {\,\lesssim\hiddensubH}
\mathcommand\consql{\,\lesssim\hiddensubCONS}
\mathcommand\hl    {\,<       \hiddensubH}
\mathcommand\hleq  {\,\leq    \hiddensubH}
\mathcommand\consl {\,<       \hiddensubCONS}
\mathcommand\conseq{\,\approx \hiddensubCONS}

\newcommand\cons {{\rm cons}}
\mathcommand\sigconsV{\sig/\cons/\V}
\mathcommand\sigconsR{\sig/\cons/\R}
\mathcommand\primesigconsV{\sig'\!/\cons'\!/\V'}
\mathcommand\primesigconsR{\sig'\!/\cons'\!/\R'}
\newcommand\dunno{{\rm sc}}
\newcommand\DUNNO{{\rm SC}}
\mathcommand\SIGCONS   {\{\SIG,\CONS\}}
\mathcommand\sigsortstimes{\SIGCONS\tight\times\sigsorts}

\mathcommand\TERMSSYM
{\mathcal{T\hspace{-.20em}E\hspace{-.08em}R\hspace{-.16em}M\hspace{-.10em}S}}
\mathapplycommand\condterms{\TERMSSYM}
\mathcommand\kurzregel{((l,r),C)}
\mathcommand\kurzregelprime{((l',r'),C')}
\mathcommand\kurzregelindex[1]{((l_{#1},r_{#1}),C_{#1})}
\mathapplycommand\lhs{\rm lhs}

\mathcommand\red{\redsimple} 

\mathcommand\lemms{L}
\mathcommand\hypos{H}
\mathcommand\goals{G}
\mathcommand\lemmsprime{\lemms'}
\mathcommand\hyposprime{\hypos'}
\mathcommand\goalsprime{\goals'}
\mathcommand\lemmsprimeprime{\lemms''}
\mathcommand\hyposprimeprime{\hypos''}
\mathcommand\goalsprimeprime{\goals''}
\mathcommand\oldtriple            {(\lemms   ,\hypos   ,\goals  )}
\mathcommand\inittriple        {(\emptyset,\emptyset,\goals  )}
\mathcommand\triplehelp[1]     {(\lemms#1,\hypos#1 ,\goals#1)}
\mathcommand\tripleprime       {\triplehelp'}
\mathcommand\triplenogoalsprime{(\lemmsprime,\hyposprime,\emptyset  )}
\mathcommand\tripleprimeprime  {\triplehelp{''}}
\mathcommand\tripleindex[1]    {\triplehelp{_{#1}}}

\mathcommand\constcong[1]{\,\,\sim_{\!_{#1}}\,}

\mathapplycommand\avail{\rm\Av ail}

\def\emph#1{\/ {\itshape#1}\/}
\newcommand\tightemph[1]{\/{\itshape#1}\/}
\newcommand\openquoteemph[1]{\ ``\hskip-0.15em{\itshape#1}\/}

\input headertheorem
\mathcommand\ident[1]{\mathsf{#1}}
\newcommand\plussymbol  {\ident{+}}
\newcommand\minussymbol {\ident{-}}
\newcommand\dividesymbol{\ident{/}}
\newcommand\timessymbol {\ident{*}}

\newcommand\set     {\ident{set}}

\newcommand\naturalssymbol{\ident{naturals}}
\newcommand\gensymsymbol{\ident{gensym}}
\mathcommand\mbpsymbol{\ident{m\hspace{-0.055em}b\hspace{-0.045em}p}}

\newcommand\csymbol     {\ident c}
\newcommand\esymbol     {\ident e}
\newcommand\fsymbol     {\ident f}
\newcommand\gsymbol     {\ident g}
\newcommand\hsymbol     {\ident h}
\newcommand\ksymbol     {\ident k}
\newcommand\psymbol     {\ident p}
\newcommand\ssymbol     {\ident s}
\newcommand\Everysymbol {\ident{Every}}
\newcommand\Permsymbol {\ident{Perm}}
\newcommand\RExistssymbol{\ident{Rexists}}
\newcommand\invertsymbol{\ident{invert}}
\newcommand\invsymbol{\ident{inv}}
\newcommand\abssymbol   {\ident{abs}}
\newcommand\cnssymbol   {\ident{cons}}
\mathcommand\cnsindexsymbol[1]{\ident{cons}_{#1}}

\newcommand\lengthsymbol{\ident{length}}
\newcommand\dlsymbol    {\ident{dl}}
\newcommand\dloncesymbol{\ident{delonce}}
\newcommand\rcsymbol    {\ident{rc}}
\newcommand\brsymbol    {\ident{br}}
\newcommand\revtailsymbol{\ident{revtail}}
\newcommand\revsymbol{\ident{rev}}
\newcommand\appendsymbol {\ident{append}}
\newcommand\zeropredicatesymbol{\ident{zerop}}
\newcommand\eqsymbol        {\ident{eq}}
\newcommand\ifthensymbol    {\mbox{\ident{If{}Then}}}
\newcommand\ifthenelsesymbol{\mbox{\ident{If{}ThenElse}}}
\mathcommand\eqindexsymbol        [1]{\eqsymbol        _{#1}}
\mathcommand\ifthenindexsymbol    [1]{\ifthensymbol    _{#1}}
\mathcommand\ifthenelseindexsymbol[1]{\ifthenelsesymbol_{#1}}
\newcommand\orsymbol    {\ident{or}}
\newcommand\andsymbol   {\ident{and}}
\newcommand\leqsymbol   {\ident{leq}}
\newcommand\lessymbol   {\ident{less}}
\newcommand\lexsymbol   {\ident{lex}}
\newcommand\acksymbol   {\ident{ack}}
\newcommand\switchsymbol{\ident{switch}}
\newcommand\swatchsymbol{\ident{swatch}}
\newcommand\diveinssymbol{\ident{div1}}
\newcommand\divzweisymbol{\ident{div2}}
\newcommand\divrestsymbol{\ident{divrest}}
\newcommand\diveinstailsymbol{\ident{div1tail}}
\newcommand\divzweitailsymbol{\ident{div2tail}}

\newcommand\turingmachinesymbol{\ident T}
\newcommand\terminatespsymbol  {\ident{terminatesp}}
\newcommand\statesymbol        {\ident{state}}
\newcommand\cmdsymbol          {\ident{cmd}}
\newcommand\nthsymbol          {\ident{nth}}
\newcommand\doublesymbol       {\ident{double}}

\newcommand\ppsymbol           {\ident{p}}
\newcommand\qpsymbol           {\ident{q}}
\newcommand\Epsymbol           {\ident{E}}
\newcommand\Ppsymbol           {\ident{P}}
\newcommand\Qpsymbol           {\ident{Q}}
\newcommand\Marriessymbol      {\ident{Marries}}
\newcommand\Lovessymbol        {\ident{Loves}}
\newcommand\StolenBysymbol     {\ident{StolenBy}}
\newcommand\Humansymbol        {\ident{Human}}
\newcommand\Evensymbol         {\ident{Even}}
\newcommand\Oddsymbol          {\ident{Odd}}
\newcommand\Primesymbol        {\ident{Prime}}
\newcommand\EveryPairsymbol   {\ident{EveryPair}}
\newcommand\Givesymbol         {\ident{Give}}
\newcommand\Fathersymbol       {\ident{Father}}
\newcommand\Elephantpsymbol    {\ident{Elephant}}
\newcommand\Flowerpsymbol    {\ident{Flower}}
\newcommand\Germanpsymbol      {\ident{German}}
\newcommand\Bicyclepsymbol     {\ident{Bicycle}}
\newcommand\Hugepsymbol        {\ident{Huge}}
\newcommand\Animalpsymbol      {\ident{Animal}}
\newcommand\Malepsymbol        {\ident{Male}}
\newcommand\Boypsymbol         {\ident{Boy}}
\newcommand\Girlpsymbol        {\ident{Girl}}
\newcommand\Femalepsymbol      {\ident{Female}}
\newcommand\Roundpsymbol       {\ident{Round}}
\newcommand\Quadrangularpsymbol{\ident{Quadrangular}}
\newcommand\Metpsymbol         {\ident{Met}}
\newcommand\Kissedpsymbol      {\ident{Kissed}}
\newcommand\Bishopsymbol       {\ident{Bishop}}
\newcommand\mindexsymbol[1]{\existsvari w{#1}}

\newcommand\nonnegpsymbol      {\ident{nonnegp}}
\newcommand\wellsymbol         {\ident{well}}
\newcommand\welltailsymbol     {\ident{welltail}}
\newcommand\varsymbol          {\ident{var}}
\newcommand\aritysymbol        {\ident{arity}}

\newcommand\whilesymbol        {\ident{while}}

\newcommand\nullsymbol         {\ident{null}}
\newcommand\hdsymbol           {\ident{hd}}
\newcommand\tlsymbol           {\ident{tl}}
\newcommand\insymbol           {\ident{in}}
\newcommand\applysymbol        {\ident{app}}
\newcommand\termsymbol         {\ident{term}}
\mathcommand\tightim{\longrightarrow}
\mathcommand\im{\ \tightim\ }
\mathcommand\rs{\:\rulesugar\:\:}
\mathcommand\rulesugar{\longleftarrow}

\mathcommand\doublepp[1]      {\doublesymbol   \beginargs{#1}\allargs}
\mathcommand\aritypp[1]      {\aritysymbol   \beginargs{#1}\allargs}
\mathcommand\lengthpp[1]      {\lengthsymbol   \beginargs{#1}\allargs}
\mathcommand\wellpp[1]      {\wellsymbol   \beginargs{#1}\allargs}
\mathcommand\welltailpp[1]      {\welltailsymbol   \beginargs{#1}\allargs}
\mathcommand\varpp[1]      {\varsymbol   \beginargs{#1}\allargs}
\mathcommand\divrestpp[2]    {\divrestsymbol\beginargs{#1}\separgs{#2}\allargs}
\mathcommand\diveinspp[2]    {\diveinssymbol\beginargs{#1}\separgs{#2}\allargs}
\mathcommand\divzweipp[3]    {\divzweisymbol\beginargs{#1}\separgs{#2}
\separgs{#3}\allargs}
\mathcommand\diveinstailpp[4]    {\diveinstailsymbol\beginargs{#1}\separgs{#2}
\separgs{#3}\separgs{#4}\allargs}
\mathcommand\divzweitailpp[6]    {\divzweitailsymbol\beginargs{#1}\separgs{#2}
\separgs{#3}\separgs{#4}\separgs{#5}\separgs{#6}\allargs}
\mathcommand\mbppp[2]         {\mbpsymbol   \beginargs{#1}\separgs{#2}\allargs}
\mathcommand\revpp[1]     
{\revsymbol\beginargs{#1}\allargs}
\mathcommand\revppi[2]     
{\revsymbol^{#1}\beginargs{#2}\allargs}
\mathcommand\revtailpp[2]     
{\revtailsymbol\beginargs{#1}\separgs{#2}\allargs}
\mathcommand\revtailppi[3]
{\revtailsymbol^{#1}\beginargs{#2}\separgs{#3}\allargs}
\mathcommand\Permpp[2]     
{\Permsymbol\beginargs{#1}\separgs{#2}\allargs}
\mathcommand\Permppi[3]
{\Permsymbol^{#1}\beginargs{#2}\separgs{#3}\allargs}
\mathcommand\appendpp[2]      
{\appendsymbol \beginargs{#1}\separgs{#2}\allargs}
\mathcommand\appendppi[3]      
{\appendsymbol^{#1}\beginargs{#2}\separgs{#3}\allargs}
\mathcommand\Everypp[2]      
{\Everysymbol \beginargs{#1}\separgs{#2}\allargs}
\mathcommand\RExistspp[1]      
{\RExistssymbol \beginargs{#1}\allargs}
\mathcommand\appendlongpp[2]      
{\appendsymbol\left(\begin{array}{@{}l@{}}{#1}\separgs\\{#2}\end{array}\right)}
\mathcommand\cnspp[2]         {\cnssymbol   \beginargs{#1}\separgs{#2}\allargs}
\mathcommand\cnsppi[3]       {\cnssymbol^{#1}\beginargs{#2}\separgs{#3}\allargs}
\mathcommand\cnsindexpp[3]
{\cnsindexsymbol{#1}\beginargs{#2}\separgs{#3}\allargs}
\mathcommand\dlpp[2]          {\dlsymbol    \beginargs{#1}\separgs{#2}\allargs}
\mathcommand\dloncepp[2]      {\dloncesymbol\beginargs{#1}\separgs{#2}\allargs}
\mathcommand\dlonceppi[3]{\dloncesymbol^{#1}\beginargs{#2}\separgs{#3}\allargs}
\mathcommand\rcpp[2]          {\rcsymbol    \beginargs{#1}\separgs{#2}\allargs}
\mathcommand\brpp[2]          {\brsymbol    \beginargs{#1}\separgs{#2}\allargs}
\mathcommand\orpp[2]          {\orsymbol    \beginargs{#1}\separgs{#2}\allargs}
\mathcommand\andpp[2]         {\andsymbol   \beginargs{#1}\separgs{#2}\allargs}
\mathcommand\shortcnspp[2]    {\csymbol     \beginargs{#1}\separgs{#2}\allargs}
\mathcommand\tightshortcnspp[2]
{\csymbol\beginargs{#1}\tightsepargs{#2}\allargs}
\mathcommand\spp[1]           {\ssymbol     \beginargs{#1}\allargs}
\mathcommand\sppiterated[2]   {\ssymbol^{#1}\beginargs{#2}\allargs}
\mathcommand\ppp[1]           {\psymbol     \beginargs{#1}\allargs}
\mathcommand\pppiterated[2]   {\psymbol^{#1}\beginargs{#2}\allargs}
\mathcommand\zeropp           {\ident 0}
\mathcommand\Julietpp         {\ident{Juliet}}
\mathcommand\Romeopp          {\ident{Romeo}}
\mathcommand\Ipp              {\ident I}
\mathcommand\onepp            {\ident1}
\mathcommand\twopp            {\ident2}
\mathcommand\threepp          {\ident3}
\mathcommand\invertpp[1]      {\invertsymbol\beginargs{#1}\allargs}
\mathcommand\invpp[1]         {\invsymbol\beginargs{#1}\allargs}
\mathcommand\abspp[1]         {\abssymbol\beginargs{#1}\allargs}
\mathcommand\naturalspp[1]    {\naturalssymbol\beginargs{#1}\allargs}
\mathcommand\gensympp[1]      {\gensymsymbol\beginargs{#1}\allargs}
\mathcommand\nilpp            {\ident{nil}}
\mathcommand\falsepp          {\ident{false}}
\mathcommand\truepp           {\ident{true}}
\mathcommand\FALSEpp          {\ident{FALSE}}
\mathcommand\TRUEpp           {\ident{TRUE}}
\mathcommand\weirdppp         {\ident{weirdp}}
\mathcommand\ambigppp         {\ident{ambigp}}
\mathcommand\zeropredicatepp[1]{\zeropredicatesymbol\beginargs{#1}\allargs}
\mathcommand\cppeins       [1]{\csymbol     \beginargs{#1}\allargs}
\mathcommand\cppzwei       [2]{\csymbol\beginargs{#1}\separgs{#2}\allargs}
\mathcommand\eppeins       [1]{\esymbol     \beginargs{#1}\allargs}
\mathcommand\fppeins       [1]{\fsymbol     \beginargs{#1}\allargs}
\mathcommand\fppeinsindex  [2]{\fsymbol_{#1}\beginargs{#2}\allargs}
\mathcommand\fppeinsiterated[2]{\fsymbol^{#1}\beginargs{#2}\allargs}
\mathcommand\gppeins       [1]{\gsymbol     \beginargs{#1}\allargs}
\mathcommand\gppzwei       [2]{\gsymbol     \beginargs{#1}\separgs{#2}\allargs}
\mathcommand\hppeins       [1]{\hsymbol     \beginargs{#1}\allargs}
\mathcommand\kppeins       [1]{\ksymbol     \beginargs{#1}\allargs}
\mathcommand\appzero          {\ident a}
\mathcommand\bppzero          {\ident b}
\mathcommand\cppzero          {\ident c}
\mathcommand\dppzero          {\ident d}
\mathcommand\eppzero          {\ident e}
\mathcommand\eqindexpp[3]{\eqindexsymbol{#1}\beginargs{#2}\separgs{#3}\allargs}
\mathcommand\ifthenindexpp
[3]{\ifthenindexsymbol{#1}\beginargs{#2}\separgs{#3}\allargs}
\mathcommand\ifthenelseindexpp
[4]{\ifthenelseindexsymbol{#1}\beginargs{#2}\separgs{#3}\separgs{#4}\allargs}
\mathcommand\eqpp[2]{\eqsymbol\beginargs{#1}\separgs{#2}\allargs}
\mathcommand\leqpp[2]{\leqsymbol\beginargs{#1}\separgs{#2}\allargs}
\mathcommand\lespp[2]{\lessymbol\beginargs{#1}\separgs{#2}\allargs}
\mathcommand\lexpp[3]{\lexsymbol\beginargs{#1}\separgs{#2}\separgs{#3}\allargs}
\mathcommand\ackpp[2]{\acksymbol\beginargs{#1}\separgs{#2}\allargs}
\mathcommand\switchpp[1]{\switchsymbol\beginargs{#1}\allargs}
\mathcommand\swatchpp[1]{\swatchsymbol\beginargs{#1}\allargs}
\mathcommand\whilepp[2]{\whilesymbol\beginargs{#1}\separgs{#2}\allargs}
\mathcommand\nullpp[1]{\nullsymbol\beginargs{#1}\allargs}
\mathcommand\nullppiterated[2]{\nullsymbol^{#1}\beginargs{#2}\allargs}
\mathcommand\hdpp[1]{\hdsymbol\beginargs{#1}\allargs}
\mathcommand\hdppiterated[2]{\hdsymbol^{#1}\beginargs{#2}\allargs}
\mathcommand\tlpp[1]{\tlsymbol\beginargs{#1}\allargs}
\mathcommand\tlppiterated[2]{\tlsymbol^{#1}\beginargs{#2}\allargs}
\mathcommand\inpp[2]{\insymbol\beginargs{#1}\separgs{#2}\allargs}
\mathcommand\inppiterated[3]{\insymbol^{#1}\beginargs{#2}\separgs{#3}\allargs}
\mathcommand\applypp[2]{\applysymbol\beginargs{#1}\separgs{#2}\allargs}
\mathcommand\applyppiterated
[3]{\applysymbol^{#1}\beginargs{#2}\separgs{#3}\allargs}
\mathcommand\termpp[2]{\termsymbol\beginargs{#1}\separgs{#2}\allargs}
\mathcommand\setpp[1]{\set\beginargs{#1}\allargs}

\mathcommand\Tpp[6]{\turingmachinesymbol\beginargs{#1}\separgs{#2}\separgs
{#3}\separgs{#4}\separgs{#5}\separgs{#6}\allargs}
\mathcommand\Tppseven[7]{\turingmachinesymbol\beginargs{#1}\separgs{#2}\separgs
{#3}\separgs{#4}\separgs{#5}\separgs{#6}\separgs{#7}\allargs}
\mathcommand\foreverppp[6]{\ident{foreverp}\beginargs{#1}\separgs{#2}\separgs
{#3}\separgs{#4}\separgs{#5}\separgs{#6}\allargs}
\mathcommand\terminatesppp[6]{\terminatespsymbol\beginargs{#1}\separgs
{#2}\separgs{#3}\separgs{#4}\separgs{#5}\separgs{#6}\allargs}
\mathcommand\terminatespppone[1]{\terminatespsymbol \beginargs{#1}\allargs}
\mathcommand\statepp
[3]{\statesymbol\beginargs{#1}\separgs{#2}\separgs{#3}\allargs}
\mathcommand\tightstatepp
[3]{\statesymbol\beginargs{#1}\tightsepargs{#2}\tightsepargs{#3}\allargs}
\mathcommand\cmdpp
[3]{\cmdsymbol  \beginargs{#1}\separgs{#2}\separgs{#3}\allargs}
\mathcommand\tightcmdpp
[3]{\cmdsymbol  \beginargs{#1}\tightsepargs{#2}\tightsepargs{#3}\allargs}
\mathcommand\stoppp           {\ident{stop}}
\mathcommand\leftpp           {\ident{left}}
\mathcommand\rightpp          {\ident{right}}
\mathcommand\nthpp         [2]{\nthsymbol  \beginargs{#1}\separgs{#2}\allargs}
\mathcommand\pppp          [1]{\ppsymbol\beginargs{#1}            \allargs}
\mathcommand\qppp          [2]{\qpsymbol\beginargs{#1}\separgs{#2}\allargs}
\mathcommand\Eppp          [1]{\Epsymbol\beginargs{#1}            \allargs}
\mathcommand\Epppzwei      [2]{\Epsymbol\beginargs{#1}\separgs{#2}\allargs}
\mathcommand\Pppp          [1]{\Ppsymbol\beginargs{#1}            \allargs}
\mathcommand\Ppppdrei      
[3]{\Ppsymbol\beginargs{#1}\separgs{#2}\separgs{#3}\allargs}
\mathcommand\Ppppvier
[4]{\Ppsymbol\beginargs{#1}\separgs{#2}\separgs{#3}\separgs{#4}\allargs}
\mathcommand\Qppp          [2]{\Qpsymbol\beginargs{#1}\separgs{#2}\allargs}
\mathcommand\Qpppeins      [1]{\Qpsymbol\beginargs{#1}\allargs}
\mathcommand\Qpppdrei      
[3]{\Qpsymbol\beginargs{#1}\separgs{#2}\separgs{#3}\allargs}
\mathcommand\Fatherpp      [2]{\Fathersymbol\beginargs{#1}\separgs{#2}\allargs}
\mathcommand\Marriespp     [2]{\Marriessymbol\beginargs{#1}\separgs{#2}\allargs}
\mathcommand\Lovespp       [2]{\Lovessymbol\beginargs{#1}\separgs{#2}\allargs}
\mathcommand\StolenBypp    [2]
{\StolenBysymbol\beginargs{#1}\separgs{#2}\allargs}
\mathcommand\Humanpp       [1]{\Humansymbol\beginargs{#1}\allargs}
\mathcommand\Evenpp        [1]{\Evensymbol\beginargs{#1}\allargs}
\mathcommand\Evenppi       [2]{\Evensymbol^{#1}\beginargs{#2}\allargs}
\mathcommand\Oddpp         [1]{\Oddsymbol\beginargs{#1}\allargs}
\mathcommand\Primepp       [1]{\Primesymbol\beginargs{#1}\allargs}
\mathcommand\EveryPairpp  [2]{\EveryPairsymbol\beginargs{#1}\separgs
{#2}\allargs}
\mathcommand\mindexppeins  [2]{\mindexsymbol{#1}\beginargs{#2}\allargs}
\mathcommand\Givepp        [3]{\Givesymbol
\beginargs{#1}\separgs{#2}\separgs{#3}\allargs}
\mathcommand\mindexppzwei  [3]{\mindexsymbol
{#1}\beginargs{#2}\separgs{#3}\allargs}
\mathcommand\mindexppdrei  [4]{\mindexsymbol
{#1}\beginargs{#2}\separgs{#3}\separgs{#4}\allargs}

\mathcommand\nonnegppp     [1]{\nonnegpsymbol\beginargs{#1}\allargs}

\mathcommand\anonymouscsymbol{c}
\mathcommand\anonymouscindexsymbol[1]{\anonymouscsymbol_{#1}}
\mathcommand\anonymousfsymbol{f}
\mathcommand\anonymouscpp
[2]{\anonymouscsymbol\beginargs{#1}\separgs\ldots\separgs{#2}\allargs}
\mathcommand\anonymouscindexpp
[3]{\anonymouscindexsymbol{#1}\beginargs{#2}\separgs\ldots\separgs{#3}\allargs}
\mathcommand\anonymousfpp
[2]{\anonymousfsymbol\beginargs{#1}\separgs\ldots\separgs{#2}\allargs}
\mathcommand\coerceindexpp[3]{[#3]_{#1}^{#2}}

\mathcommand\Elephantppp    [1]{\Elephantpsymbol\beginargs{#1}\allargs}
\mathcommand\Flowerppp      [1]{\Flowerpsymbol  \beginargs{#1}\allargs}
\mathcommand\Bicycleppp     [1]{\Bicyclepsymbol \beginargs{#1}\allargs}
\mathcommand\Germanppp      [1]{\Germanpsymbol  \beginargs{#1}\allargs}
\mathcommand\Hugeppp        [1]{\Hugepsymbol    \beginargs{#1}\allargs}
\mathcommand\Animalppp      [1]{\Animalpsymbol  \beginargs{#1}\allargs}
\mathcommand\Maleppp        [1]{\Malepsymbol    \beginargs{#1}\allargs}
\mathcommand\Boyppp         [1]{\Boypsymbol     \beginargs{#1}\allargs}
\mathcommand\Girlppp        [1]{\Girlpsymbol    \beginargs{#1}\allargs}
\mathcommand\Femaleppp      [1]{\Femalepsymbol  \beginargs{#1}\allargs}
\mathcommand\Roundppp       [1]{\Roundpsymbol   \beginargs{#1}\allargs}
\mathcommand\Bishoppp       [1]{\Bishopsymbol   \beginargs{#1}\allargs}
\mathcommand\Quadrangularppp[1]{\Quadrangularpsymbol  \beginargs{#1}\allargs}
\mathcommand\Kissedppp[2]{\Kissedpsymbol\beginargs{#1}\separgs{#2}\allargs}
\mathcommand\Metppp[2]   {\Metpsymbol   \beginargs{#1}\separgs{#2}\allargs}

\newcommand\bound     {{\rm bound}}
\newcommand\free      {{\rm free}}

\mathcommand\Vtripleindex[3]{\V\!_{{#1},\,{#2},\,{#3}}}
\mathcommand\Vdoubleindex[2]{\V\!_{{#1},\,{#2}}}
\mathcommand\Vsingleindex[1]{\V\!_{{#1}}}

\mathcommand\Erel[1]{\Gammaoffont\!_{#1}}
\mathcommand\Urel[1]{\Deltaoffont_{#1}}

\mathcommand\theRprimefromstrongtoweak{
  \inparenthesesinlinetight{
     \domres\id{\Vwall\cup\Vsome\setminus\RAN\varsigma}
     \nottight{\nottight\uplus}
     \reverserelation\varsigma
  }
  \nottight{\circ}
  \ranres
    {\transclosureinline R}
    {\Vwall\cup\Vsome\setminus\RAN\varsigma}
  \nottight{\nottight{\nottight{\uplus}}}
  \Vsome\tighttimes\Vsall
}

\mathcommand\deltaminus{\delta^-}
\mathcommand\deltaplus{\delta^+}
\mathcommand\deltaplusplus{\delta^{+^+}}
\mathcommand\deltastar{\delta^*}
\mathcommand\deltastarstar{\delta^{*^*}}

\mathcommand\Vall     {\Vsingleindex\indexdelta         }
\mathcommand\Vwall    {\Vsingleindex\indexdeltaminu     }
\mathcommand\Vsall    {\Vsingleindex\indexdeltaplus     }
\mathcommand\Vgsome   {\Vsingleindex\indexgammaplus     }
\mathcommand\Vsome    {\Vsingleindex\indexgamma         }
\mathcommand\Vfree    {\Vsingleindex\indexfree          }
\mathcommand\Vbound   {\Vsingleindex\indexbound         }
\mathcommand\Vsomesall{\Vsingleindex\indexgammadeltaplus}

\mathapplycommand\VARall      {\VARsingleindex\indexdelta         }
\mathapplycommand\VARwall     {\VARsingleindex\indexdeltaminu     }
\mathapplycommand\VARsall     {\VARsingleindex\indexdeltaplus     }
\mathapplycommand\VARgsome    {\VARsingleindex\indexgammaplus     }
\mathapplycommand\VARsome     {\VARsingleindex\indexgamma         }
\mathapplycommand\VARfree     {\VARsingleindex\indexfree          }
\mathapplycommand\VARbound    {\VARsingleindex\indexbound         }
\mathapplycommand\VARsomesall {\VARsingleindex\indexgammadeltaplus}
\mathcommand\displayVARsall[1]{\VARsingleindex\indexdeltaplus
\!\!\!\:\left(\begin{array}{@{}c@{}}#1\end{array}\right)}

\mathcommand\rigidvari     [2]{#1_{#2}^\indexgammadeltaplus}
\mathcommand\existsvari    [2]{#1_{#2}^\indexgamma    }
\mathcommand\forallvari    [2]{#1_{#2}^\indexdelta    }
\mathcommand\freevari      [2]{#1_{#2}^\indexfree     }
\mathcommand\wforallvari   [2]{#1_{#2}^\indexdeltaminu}
\mathcommand\sforallvari   [2]{#1_{#2}^\indexdeltaplus}
\mathcommand\gexistsvari   [2]{#1_{#2}^\indexgammaplus}
\mathcommand\boundvari     [2]{#1_{#2}}
\mathcommand\vari          [2]{#1_{#2}}
\mathcommand\wforallvarilow[2]{#1_{#2}^
{\raisebox{-.82ex}{\math\indexdeltaminu}}}

\newcommand\indexhelper[1]{{\scriptscriptstyle#1\:\!\!}}
\newcommand\indexdeltaplus
{\indexhelper{\delta^{\raisebox{-.17ex}{\fvesf\hskip-0.14em +}}}}
\newcommand\indexdeltaminu
{\indexhelper{\delta^{\mbox{\fvesf\hskip-0.14em\rule[.2ex]{.7em}{.15ex}}}}}
\newcommand\indexgammaplus
{\indexhelper{\gamma^{\mbox{\fvesf\hskip-0.14em +}}}}
\newcommand\indexgammadeltaplus
{\indexhelper{\gamma\delta^{\raisebox{-.17ex}{\fvesf\hskip-0.14em +}}}}

\newcommand\indexdelta{\indexhelper\delta}
\newcommand\indexgamma{\indexhelper\gamma}
\newcommand\indexfree
{{\scriptscriptstyle\free}}
\newcommand\indexbound
{{\scriptscriptstyle\bound}}

\newcommand\Wellfsymb{\ident{Wellf}}
\mathapplycommand\Wellfpp{\Wellfsymb}

\mathcommand\beginargs{(}
\mathcommand\allargs  {)}
\mathcommand\separgs  {,\,}
\mathcommand\tightsepargs{,}

\mathcommand\minusppnoparentheses  [2]{{#1}\,\minussymbol\,{#2}}
\mathcommand\tightminusppnoparentheses  [2]{{#1}\minussymbol{#2}}
\mathcommand\divideppnoparentheses [2]{{#1}\,\dividesymbol\,{#2}}
\mathcommand\plusppnoparentheses   [2]{{#1}\,\plussymbol \,{#2}}
\mathcommand\plusppnoparenthesesi  [3]{{#2}\,\plussymbol^{#1}\,{#3}}
\mathcommand\tightplusppnoparentheses   [2]{{#1}\plussymbol{#2}}
\mathcommand\timesppnoparentheses  [2]{{#1}\,\timessymbol\,{#2}}
\mathcommand\undppnoparentheses    [2]{{#1}\und            {#2}}
\mathcommand\oderppnoparentheses   [2]{{#1}\oder           {#2}}
\mathcommand\impliesppnoparentheses[2]{{#1}\implies        {#2}}
\mathcommand\leqinfixppnoparentheses[2]{{#1}\,\tight\leq\,{#2}}
\mathcommand\geqinfixppnoparentheses[2]{{#1}\,\tight\geq\,{#2}}
\mathcommand\dividepp [2]{(\divideppnoparentheses {#1}{#2})}
\mathcommand\minuspp  [2]{(\minusppnoparentheses  {#1}{#2})}
\mathcommand\pluspp   [2]{(\plusppnoparentheses   {#1}{#2})}
\mathcommand\timespp  [2]{(\timesppnoparentheses  {#1}{#2})}
\mathcommand\undpp    [2]{(\undppnoparentheses    {#1}{#2})}
\mathcommand\oderpp   [2]{(\oderppnoparentheses   {#1}{#2})}
\mathcommand\impliespp[2]{(\impliesppnoparentheses{#1}{#2})}

\newcommand\englishtextelevenone
{\englishtexteleventwo{Not only the notorious golden mountain is of gold, 
 but also\\the}.}
\newcommand\englishtexteleventwo[1]
{#1 round quadrangle is just as certainly round as it is quadrangular}

\newcommand\frenchtextone{%
Je fus longtemps sans pouvoir appliquer 
ma m\'ethode aux questions affirmatives, 
parce que le tour et le biais pour y venir est beaucoup plus 
malais\'e que celui dont je me sers aux n\'egatives. 
De sorte que, lorsqu'il me fallut d\'emontrer que\emph
{tout nombre premier, qui surpasse de l'unit\'e un multiple de 4, 
 est compos\'e de deux quarr\'es}, je me trouvai en belle peine.
Mais enfin une m\'editation diverses fois r\'eit\'er\'ee me donna
les lumi\`eres qui me manquoient, et les questions affirmatives pass\`erent 
par ma m\'ethode, \`a l'aide de quelques nouveaux principes qu'il
y fallut joindre par n\'ecessit\'e.}
\newcommand\frenchtextonecitation{\cite[\Vol\,II, \p\,432]{fermat-oeuvres}}

\newcommand\englishtextonehundredandthirteenquotation
{our translation}

\mathcommand\termsofdepthnovars[1]{{\mathcal T}_{#1}}

\newcommand\englishtextninehundred
{Thus, in mathematics, we have no reasons to assume any meaning of
 `existence' that would be fundamentally different from that of 
 `the validity of axiomatic relations\closesinglequotefullstopnospace}

\catcode`\@=11
\input ENDNOTES.sty
\catcode`\@=12
\def\enotesize{\normalsize}
\def\endnoteendsep{\vspace{5.0\topsep}}
\let\footnote=\endnote
\input xy
\xyoption{ps}
\xyoption{all}
\xyoption{dvips}
\usepackage{named}
\def\citep{\cite}
\def\citet#1{\citeauthor{#1} \shortcite{#1}}
\newcommand\startcite{{\raise.2ex\hbox{[}}}
\newcommand\stopcite {\raise.2ex\hbox{]}}
\newcommand\citehelper[1]{\startcite #1\stopcite}

\newcommand\makeaciteofthree[3]
{\citehelper{\citeauthor{#1}, \citeyear{#1}; \citeyear{#2}; \citeyear{#3}}}

\date
{\vspace*{-2ex}\footnotesize\begin{tabular}{c} 
 Submitted: 2006, \Aug\,20. \ 
 First Print Edition: 2006, \Nov\,2.
 \\Thoroughly Revised Second Edition: 2010, \Dec\,14
 \\\end{tabular}}
\catcode`\@=11
\renewcommand\tableofcontents{%
    \section*{\contentsname
        \@mkboth{%
           \MakeUppercase\contentsname}{\MakeUppercase\contentsname}}%
    \vskip .2ex
    \@starttoc{toc}%
    }
\catcode`\@=12
\renewcommand\namefont{\sc}%
\newcommand\frenchtexttwosubtextone
{qui auroit \frenchtexttwosubtextthree}
\newcommand\frenchtexttwosubtexttwo{moindre que le}
\newcommand\frenchtexttwosubtextthree{la m\^eme propri\'et\'e}
\newcommand\frenchtexttwo{%
S'il y avoit aucun triangle rectangle en nombres entiers qui e\^ut son 
aire \'egale a un quarr\'e, 
il y auroit un autre triangle moindre que
celui-l\`a, \frenchtexttwosubtextone.
S'il y en avoit un second,
\frenchtexttwosubtexttwo\ premier, 
qui e\^ut \frenchtexttwosubtextthree,
il y en auroit, 
par un pareil raisonnement, 
un troisi\`eme, 
\frenchtexttwosubtexttwo\ second, 
\frenchtexttwosubtextone, 
et enfin un quatri\`eme, un cinqui\`eme, \nolinebreak\etc\
\`a l'infini en descendant.%
}
\newcommand\latintextone{%
Si area trianguli esset quadratus,
darentur duo quadratoquadrati quorum differentia esset quadratus;
unde sequitur dari duo quadratos quorum et summa et differentia esset quadratus.
Datur itaque numerus,
compositus ex quadrato et duplo quadrati,
aequalis quadrato,
ea conditione ut quadrati eum componentes faciant quadratum.

Sed,
si numerus quadratus componitur ex quadrato et duplo alterius quadrati,
eius latus similiter componitur ex quadrato et duplo quadrati,
ut facillime possumus demonstrare.

Unde concludetur latus illud esse summam laterum circa 
rectum trianguli rectanguli,
et unum ex quadratis illud componentibus efficere basem,
et duplum quadratum aequari perpendiculo.}
\newcommand\latintexttwo{%
Illud itaque triangulum rectangulum conficietur a duobus quadratis quorum 
\linebreak
summa et differentia erunt quadrati.
At isti duo quadrati minores probabuntur primis quadratis primo suppositis,
quorum tam summa quam differentia faciunt quadratum:

Ergo,
si dentur duo quadrati quorum summa et differentia faciunt quadratum,
dabitur in integris summa duorum quadratorum eiusdem naturae,
priore minor.
}
\newcommand\latintextthree{%
Eodem ratiocinio dabitur et minor ista inventa per viam prioris,
et semper in infinitum minores in\-veni\-entur numeri in integris 
idem praestantes:
Quod impossibile est,
quia,
dato numero quovis integro,
non possunt dari infiniti in integris illo minores.
\par
Demonstrationem integram et fusius explicatam 
inserere margini vetat ipsius exiguitas. 
}
\newcommand\latintextten{Hanc marginis exiguitas non caperet.}
\newcommand\frenchtextoneinenglish{%
For a long time I was not able to apply my method to affirmative conjectures
because the ways and means of achieving this are much more complicated than
the ones I am used to for negative conjectures.
Such that, when I had to show that any prime number which exceeds
1 by a multiple of 4 is the sum of two squares, 
I found myself pretty much in trouble.
But finally oft-repeated meditation gave me the insight I lacked,
and affirmative questions yielded to my method with
the aid of some new principles which had to be added to it.}
\newcommand\frenchtexttwoinenglishsubtextone
{which would have the same property}
\newcommand\frenchtexttwoinenglish{%
If there were any right-angled triangle in whole numbers 
that had its area equal to a square, there would be another 
triangle smaller than that one,
\frenchtexttwoinenglishsubtextone. 
If there were a second, smaller than the first, which had the same property,
there would be, by a similar reasoning, a third, smaller than the second,
\frenchtexttwoinenglishsubtextone, 
and finally a fourth, a fifth, \etc, descending to infinity.}
\newcommand\OBS{Observation\,XLV}
\newcommand\OBSlong{Observation\,XLV on \diophantus' Problem\,XX}
\newcommand\OBSciteinswhole{\cite[\Vol\,VI, \p\,338\f]{diophantus-fermat}}
\newcommand\OBSciteinsclose{\cite[\Vol\,VI, \p\,339]{diophantus-fermat}}
\newcommand\OBScitzwei{\cite[\Vol\,I, \p\,340\f]{fermat-oeuvres}}

\begin{document}
\setcounter{tocdepth}{2}\makecover
\maketitle\notop\notop\thispagestyle{empty}\begin{abstract}%
We present the only proof of \fermatname\ 
by\emph\descenteinfinie\ that is known to exist today.
As the text of its Latin original requires active mathematical interpretation,
it is more a proof {\em sketch}\/ than a proper mathematical proof.
We discuss {\it\descenteinfinie}\/ from the 
mathematical, logical, historical, linguistic, 
and refined logic-historical points of view.
We provide the required preliminaries from number theory
and develop a self-contained proof in a modern form,
which nevertheless is intended to follow \fermat's ideas closely. 
We then annotate an English translation of \fermat's original proof with 
terms from the modern proof.
Including all important facts, 
we \nolinebreak present a concise and self-contained 
discussion of \fermat's proof sketch, 
which is easily accessible to laymen in number theory as well as
to laymen in the history of mathematics, 
and which provides new clarification of the 
Method of {\it\DescenteInfinie}\/ to the experts in these fields.
Last but not least, this \daspaper\ fills a gap regarding the 
easy accessibility of the subject.\notop\end{abstract}\renewcommand\abstractname
{R\'esum\'e}\begin{abstract}%
Nous pr\'esentons la seule preuve de Pierre Fermat par descente infinie qui
est connue aujourd'hui.
Car le texte de son origine latine exige une interpr\'etation math\'ematique
active, il est plus un croquis 
de preuve qu'une preuve math\'ematique compl\`ete. Nous discutons la descente
infinie par les points 
de vues math\'ematique, logique, historique, linguistique et par un point de
vue logique-historique raffin\'e.
Nous fournissons les pr\'eliminaires n\'ecessaires \`a partir de la th\'eorie des
nombres et nous d\'eveloppons 
une preuve en forme moderne, qui se satisfait \`a elle-m\^eme et qui suit
pourtant de pr\`es les id\'ees de Fermat.
Nous continuons par annoter une traduction anglaise de la preuve originale
de Fermat avec les termes 
de la preuve moderne. Y compris tous les faits importants, nous pr\'esentons
une discussion concise et 
coh\'erente du croquis de preuve de Fermat, qui est facilement compr\'ehensible
pour des profanes 
en th\'eorie des nombres ainsi pour des profanes en l'histoire des
math\'ematiques, et qui offre des 
explications nouvelles de la m\'ethode de descente infinie aux experts dans
ces domaines.
Finalement ce document comble une lacune en ce qui concerne la
compr\'ehensibilit\'e du sujet.\pagebreak\end{abstract}\tableofcontents\vfill\pagebreak
\section{Introduction and Motivation}\label
{section Introduction}
It seems that
---~for pedagogical as well as political reasons~---
myth has to surround the truly 
paradigmatic figures in the history of science
with fictitious association disconnected from the historical facts. \ 
\galileiname\ \galileilifetime\ is the primary example for this; \ 
\cfnlb\ \cite{against-method}, \cite{prause-galilei}. \ 
But also the most famous mathematician \fermatname\ is a subject of myth. \ 
For instance, on the one hand, 
one has tried to turn \fermat\ into a model for mankind,
\cfnlb\ \cite{fermat-saint}. \ On the other hand, 
\fermat\ was accused to be a rogue:
\begin{quote}``Actions, however, speak louder than words.
The fact that none of the many letters of \fermat\ 
which survive gives any real indication of his methods surely means that,
consciously or unconsciously,
he was very jealous, secretive, and competitive about his work,
as were all of his contemporaries.''
\getittotheright{\cite[\litsectref{1.6}, \p 11]{fermat-last-theorem}}
\end{quote}
The myth on \fermat\ even continues with his name and his life time,
for a funny collection \cf\ \cite{catherine-goldstein}, \litsectref{1}. \ 
The most famous mathematician \fermatname\ 
was born not in \nolinebreak 1601 as usually claimed, but
either in \nolinebreak 1607 or in \nolinebreak January\,1608. \ 
The \fermatname\ born in 1601 died before his stepbrother, 
our most famous \fermatname\ was born.
Our \fermat\ was a competent lawyer 
and devoted judge of the parlement of Toulouse (\tightemph\fermatsprofession),
a position which he bought and by which he was admitted the 
title\emph\fermatstitle.
Thus, \fermatnoblename\ is the address to the noble judge.
\par\yestop\noindent
The mathematician \fermatname\ 
only existed in the very rare leisure time of this most busy judge.
What would mathematics be like today if this incredible genius 
would not have put mathematics behind family, profession, social status,
and commerce?
For more up-to-date information on \fermat's life
we recommend \cite{fermatslife} instead of the better known 
\cite{fermat-career},
as the latter provides 
reliable information only on the mathematics of \fermat.
\par\yestop\noindent
All what is important for us here, is that the
field of\emph{number theory} as we know it today, was basically created by 
the mathematician \fermatname\ \fermatlifetime.
He built and improved on \diophantusname\ \diophantuslifetime,
who had looked for rational solutions of
a large number of problems in number theory. \ 
In number theory, 
\fermat\ left the classical association to geometry behind
(but was more ingenious than \vietename\ \vietelifetime)
and insisted on integer solutions for the problems. \ 
\fermat's mathematical work on number theory
seems to have taken place during his very rare leisure time in his easy chair,
from which he avoided to get up to fetch paper.
Instead, he scribbled his 
ideas into his copy of \diophantus'\emph{Arithmetic} in the 
commented bilingual Greek and Latin
edition \cite{diophantus-bachet} of \bachetname\ \bachetlifetime\@. \ 
It \nolinebreak is not surprising that these notes
---~intended for the most skilled and ingenious problem-solver \fermat\ himself
to reconstruct his findings~---
are very short and hard to understand.
In \nolinebreak his letters to contemporary mathematicians, however,
he used to be just as short or even shorter.
The reason seems to be that \fermat\ 
wanted his correspondents to do number theory on their own and to find out
how much fun it \nolinebreak is, \cfnlb\ \cite{fermat-career}.
Reporting piecemeal on his results and his methods, but hiding 
his theorems in their most general form and his proofs, 
he became a most famous but lonely mathematician.\vfill\pagebreak
\par\yestop\noindent
The Method of\emph\DescenteInfinie\ 
is the standard induction method of the working mathematician
from the ancient Greeks until today.
It got lost in the Middle Ages and was reinvented and named by \fermat,
\nolinebreak\cfnlb\ \sectref{section descente infinie}. \ 
\par\yestop\noindent
\fermat's marginal notes in his copy of \diophantus'\emph{Arithmetic} 
were published in 1670 only a few years after his death,
and in this \daspaper\
we will have a look at a text passage of \OBS\ of these
{\em Observations on \diophantus}; \ \cfnlb\ \OBSciteinswhole, \ \OBScitzwei. \ 
This passage contains the only proof of \fermat\ by
the\emph{Method of \DescenteInfinie} explicitly known today.
Already by this fact, \OBS\ is
a most important and precious piece of mathematics.
It becomes even more important by the fact that 
it paradigmatically exemplifies the Method of\emph\DescenteInfinie\ 
and exhibits this method's conceptual aspects and technical problems
in a multitude which is truly surprising for such a short text.
All in all, \fermat's \OBS\ is the primary example for 
the Method of\emph\DescenteInfinie, 
historically, conceptually, and pedagogically.
\par\yestop\noindent
As \fermat's original proof is hard to understand,
we first have to grasp the mathematical ideas implicitly expressed 
in this proof.
Note that this cognitive process is similar to the interpretation of a 
music passage from its notes in the following sense: 
If we perceive a gestalt
of the passage, this gestalt will be meaningful, but not necessarily the 
original one of the author.
After projecting our image onto the original passage,
we can then evaluate its adequacy.
When we look at a text of the \nth{17} century today,
we are very likely to interpret something into it, however, 
which \fermat's contemporaries would not have done.
Nevertheless, I \nolinebreak may hope that this paper is not infected 
by more modern number theory simply for the following reason:
I \nolinebreak did not do number theory seriously the last twenty years.
And I did not use
any further material on number theory beside \euclid's\emph{Elements}, 
but did everything on my own 
without getting out of my easy chair.
It took me a couple of days, but it was an incredible lot of fun.
This indicates that 
\fermat\ was right and his contemporaries should not have neglected 
his challenges; \ \cfnlb\ \cite{fermat-career}.
\par\yestop\noindent
My mixed motivations for writing this \daspaper\ were actually the following:
\begin{enumerate}
\item
There was no concise presentation of the subjects including all important facts
and being easily accessible to laymen in the history of mathematics.
\item\sloppy
Regarding \fermat's proof,
there was no easily comprehensible 
self-contained presentation suited for a student in 
computer science with a minor knowledge in number \mbox{theory}. \ 
This \mbox\daspaper\ should enable him to carry out a case study with
our inductive theorem proving software system \QUODLIBET; \ 
\cfnlb\ 
\cite{quodlibet-cade}, \hskip.2em
\makeaciteofthree{samoacalculemus}{samoa-phd}{jancl}, \hskip.2em
\makeaciteofthree{wirthcardinal}{zombie}{swp200601}.\item
I wanted to have fun and to 
construct a \naive\ interpretation of \fermat's proof
that has a good chance
to be more in the style of the \nth{17}\,century 
than interpretations of modern experts in number theory.
\item Moreover, with my expertise in logic and automated theorem proving,
I had to clarify some methodological aspects of {\em\DescenteInfinie}\/
and to make some minor contributions to the interpretation of \fermat's proof.
\end{enumerate}

\vfill\pagebreak

\section{{\em\DescenteInfinie}}\label{section descente infinie}
\subsection{Working Mathematician's Point of View}\label
{subsection Working Mathematician's Point of View}
In everyday mathematical practice of an advanced theoretical journal the 
frequent inductive arguments 
are hardly ever carried out explicitly. \
Instead, \hskip.1em
the proof just reads something like 
\hskip.25em ``by structural induction on \math n, \qedabbrev''\ \hskip.25em
or 
\hskip.25em``by induction on \pair x y over \math <, 
\qedabbrev\closequotecommaextraspace
expecting that the mathematically educated reader could easily expand the 
proof if in \nolinebreak doubt. \ 
In \nolinebreak contrast, 
very difficult inductive arguments, sometimes covering several 
pages, such as the proofs of \hilbert's
{\em\nth 1\,\,\math\varepsilon-theorem}, \ 
\gentzen's {\em Hauptsatz}, \ 
or confluence theorems such as the ones in 
\cite{gwrta} and \cite{wirth-jsc} \hskip.2em
still require considerable ingenuity and {\em will}\/ be
carried out! \ 
The experienced mathematician engineers his proof roughly according to
the following pattern:
\begin{quote}\howdescenteinfiniegoes\end{quote}\noindent\label{section items}%
The hard tasks of proof by mathematical induction are
\begin{description}\item[(Hypotheses Task) ]\mbox{}\\{to find the
 numerous induction hypotheses (as, \eg, 
 in the proof of \gentzen's Hauptsatz on Cut-elimination in 
 \cite{gentzen})}
{and}\item[(Induction-Ordering Task) ]\mbox{}\\{to construct 
 an\emph{induction ordering} for the proof, \ie\ 
 a \wellfounded\ ordering that satisfies
 the ordering constraints of all
 these induction hypotheses in parallel. (For instance, this was the hard
 part in the elimination of the \math\varepsilon-formulas
 in the proof of the \nth 1\,\math\varepsilon-theorem
 in \cite[\Vol\,II]{grundlagen}, \ 
 and in the proof of 
 the consistency of arithmetic by the \nlbmath\varepsilon-substitution
 method in 
 \cite{ackermann-consistency-of-arithmetic}).}
\end{description}\par\noindent
The soundness of the above
method for engineering hard induction proofs 
is easily seen when the argument is structured as a proof by contradiction,
assuming a counterexample.
For \nolinebreak
\fermat's historic reinvention of the method,
it is thus just natural that he developed 
the method itself in terms of assumed counterexamples. 
He \nolinebreak
called it\openquoteemph{\descenteinfinie\ ou ind\'efinie}\closequotefullstop
Here is this\emph{Method of \DescenteInfinie} in modern language, 
very roughly speaking:
A proposition \math\Gamma\ can be proved by\emph\descenteinfinie\ as follows:
\begin{quote}\em\howMethodofDescenteInfiniegoes\end{quote}
\vfill\pagebreak

\subsection{Logical Point of View}\yestop\noindent\thelogicalpointofview
\par\yestop\noindent
The proposition \nlbmath\Gamma\ of 
\sectref{subsection Working Mathematician's Point of View}
is represented in \inpit{\ident{N}} by \bigmaths{\forall x\stopq\app P x}. \ 
Roughly speaking, 
\mbox{a\emph{counterexample}}
for \nlbmath\Gamma\ is an instance \nlbmath{a} for which 
\bigmath{\neg\app P a}{} holds, but we should be more careful here 
because this is actually a semantical notion and not a syntactical one;
\cfnlb\ \cite{wirthcardinal}, \litsectref{2.3.2}. \ 
To treat counterexamples properly,
we need 
a logic that actually models
the mathematical process of proof search by\emph\descenteinfinie\ itself and 
directly supports it with the data structures required for a formal treatment,
and thus requires a semantical treatment of free variables.
The only such logic can be found in \cite{wirthcardinal}.
\vfill\pagebreak

\subsection{Historical Point of View}\label
{subsection Historical Point of View}
\subsubsection{Early Greek History}
Although we do not have any original Greek mathematical documents 
from the \nth 5 \nolinebreak century \BC\ and only fragments from the 
following millennium,
the first known occurrence of\emph\descenteinfinie\ in history 
seems to be the proof of the irrationality of the golden number
\bigmaths{\frac 1 2\inpit{1\tight+\sqrt 5}}{} by the Pythagorean mathematician
\hippasosname\ (Italy) in the middle of the \nth 5 century 
\BC, \cf\ \cite{hippa}.
This proof is carried out geometrically in a pentagram,
where the golden number gives the proportion
of the length of a line to the length of the side 
of the enclosing pentagon:\halftop
\initial{\raisebox{-3.4ex}{\includegraphics
[scale=1,bb=201 467 232 497]{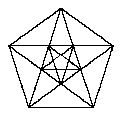}}}
Under the assumption that this proportion is given by 
\bigmaths{m:n}{} with natural numbers \nlbmath m and \nlbmath n, 
it can be shown that the proportion of the length
of a line of a new pentagram drawn inside the  
inscribed pentagon to the length of the side of this pentagon
is \bigmaths{m\tight-n:2n\tight-m}, with 
\bigmaths{0\prec m\tight-n\prec m}, and so forth since the new inscribed
pentagram is similar to the original one. \ 
A myth says that the gods drowned \hippasos\ in the sea,
as a punishment for destroying the Pythagoreans' belief 
that everything is given by positive rational numbers;
and this even with the pentagram, which was  
the Pythagoreans' sign of recognition amongst
themselves.
The resulting confusion seems to have 
been one of the reasons for the ancient Greek 
culture to shift interest in mathematics from theorems to proofs.

\subsubsection{\euclid's Elements}
\subsubsubsection{Proof by Generalizable Example}
In the famous collection ``\hskip-.18em\tightemph{Elements}'' 
of \euclidname\ \euclidlifetime,
we find several occurrences of\emph\descenteinfinie. 
In the Elements, the verbalization of a proof 
by\emph\descenteinfinie\ has the form of 
a\emph{generalizable example}
in the sense that a 
special concrete counterexample is 
considered
---~instead of an arbitrary one~---
but the existence of a smaller
counterexample is actually shown independently of this special choice.
Similarly, the induction step of a structural induction may also be presented
in the form of a generalizable example.
Such proofs via a generalizable example are called\emph{quasi-general} 
in \cite{freudenthal}.
We would not accept a quasi-general proof 
as a proper proof from our students today because 
the\emph{explicit} knowledge and the\emph{explicit} verbalization
of methods of mathematical induction
have become standard during the last centuries.
And we may ask why the Elements proceed by generalizable examples.
For this question it is interesting to see that already 
in a text of \platoname\ \platolifetime\ (Athens) 
we find 
a proof by structural induction with a proper verbalization 
of a general induction step without resorting to generalizable examples,
\cfnlb\ \cite{plato-induction}.
As the theorem of this proof is mathematically trivial 
(``\math{n\tight+1} terms in a list have \nlbmath{n} contacts''), 
the intention of this proof seems to be the explicit demonstration of
the activity of structural induction itself,
though no instance of the 
\axiomofstructuralinduction\ \nlbmath{\inpit{\ident{S}}} is explicitly 
mentioned.
Moreover, the verbalization of a variable number and even the comprehension
of a non-concrete example and a general induction proof seems to have 
been a challenge for an ancient Greek student; \
\cfnlb\ \cite[\p\,279\ff]{unguru-one}. \ 
Thus, 
the presentation of induction proofs via generalizable examples 
in the Elements may well have had pedagogical reasons.
\vfill\pagebreak

\yestop\noindent
Let us have a look at two proofs from the Elements.

\subsubsubsection{Proof by\emph\DescenteInfinie}
In \cite[\Vol\,VII, \litpropref{31}]{elements}, \hskip.25em
we find the following proof by {\em\descenteinfinie}:
\begin{quote}
\underline{\litpropref{VII.31}:} \ 
Any composite number is measured by some prime number.
\par\noindent\underline{Proof of \litpropref{VII.31}:} \
Let \nlbmath A be a composite number.
I say that \nlbmath A is measured by some prime number.
Since \nlbmath A is composite, therefore some number \nlbmath B measures it. 
Now, if \nlbmath B is prime, then that which was proposed is done. 
But if it is composite, some number measures it. 
Let a number \nlbmath C measure it. 
Then, since \nlbmath C measures \nlbmath B, and \nlbmath B measures \nlbmath A, 
therefore \nlbmath C also measures \nlbmath A. 
And, if \nlbmath C is prime, then that which was proposed is done. 
But if it is composite, some number measures it. 
Thus, if the investigation is continued in this way, 
then some prime number will be found which measures the number before it, 
which also measures \nlbmath A. 
If it is not found, 
then an infinite sequence of numbers measures the number \nlbmath A, 
each of which is less than the other, which is impossible in numbers.
Therefore some prime number will be found which 
measures the one before it, which also measures \nlbmath A.
\QED{\litpropref{VII.31}}\end{quote}

\yestop
\subsubsubsection{Proof by Structural Induction}
Taking into account that the ancient Greeks were not familiar with
an actually infinite set of natural numbers, \hskip.2em
in accordance with \cite{freudenthal} \hskip.2em
I consider the proof of \litpropref{IX.8} of the Elements to be
obviously a proof by structural induction, \hskip.2em
whereas \cite{unguru-one} rejects this opinion and
\cite{plato-induction} even claims that there are no proofs by 
structural induction in \euclid's Elements at all. \
Thus, \hskip.2em
let us have a look at this proof to give the reader a chance to 
judge on his own.
\begin{quote}
\underline{\litpropref{IX.8}:} \ 
If as many numbers as we please beginning from a unit are in continued 
proportion, then the third from the unit is square 
as are also all those which successively leave out one, 
and the fourth is cubic 
as are also all those which leave out two, 
and the seventh is both cubic and square 
as are also all those which leave out five.
\par\noindent\underline{Proof of \litpropref{IX.8}:} \
Let there be as many numbers as we please, 
\math A, \math B, \math C, \math D, \math E, and \math F, 
beginning from a unit and in continued proportion. 
I say that \nlbmath B, the third from the unit, 
is square as are all those which leave out one; 
\math C, the fourth, is cubic as are all those which leave out two; 
and \math F, the seventh, 
is both cubic and square as are all those which leave out five. 
Since the unit is to \nlbmath A as \math A is to \math B, 
therefore the unit measures the number \math A the same number of times that 
\math A measures \math B. 
But the unit measures the number \math A according to the units in it, 
therefore \math A also measures \math B according to the units in \math A. 
Therefore \math A multiplied by itself makes \math B, 
therefore \math B is square. 
And, since \math B, \math C, and \math D are in continued proportion, 
and \math B is square, therefore \math D is also square. 
For the same reason \math F is also square. 
Similarly we can prove that all those which leave out one are square.
I say next that \math C, the fourth from the unit, is cubic 
as are also all those which leave out two.
Since the unit is to \math A as \math B is to \math C, 
therefore the unit measures the number \math A the same number of times that 
\math B measures \math C. 
But the unit measures the number \math A according to the units in \math A, 
therefore \math B also measures \math C according to the units in \math A. 
Therefore \math A multiplied by \nlbmath B makes \math C. 
Since then \math A multiplied by itself makes \math B, 
and multiplied by \nlbmath B makes \math C, therefore \math C is cubic.
And, since \math C, \math D, \math E, and \math F are in continued proportion,
and \math C is cubic, therefore \math F is also cubic. 
But it was also proved square, 
therefore the seventh from the unit is both cubic and square. 
Similarly we can prove that all the numbers which leave out 
five are also both cubic and square.
\QED{\litpropref{IX.8}}
\end{quote}

\halftop
\subsubsection{Recovering from the Dark Middle Ages}
After \euclid, in the following eighteen centuries until \fermat,
I do not know of\emph\descenteinfinie\
(except that \euclid's Elements where copied again and again),
but of\emph{structural induction} only.
Structural induction was known to the Muslim mathematicians around
the year 1000 and occurs in a Hebrew book of \gersonname\ \gersonlifetime\
(Orange and \Avignon) in 1321, \cf\ \cite{katz-history}. \
\pascalname\ \pascallifetime\ (Paris) knew structural induction from
``\hskip-.18em\tightemph{Arithmeticorum Libri Duo}'' 
of \maurolycusname\ (\maurolycuslatin) 
\maurolycuslifetime\ (\maurolycuslifeplace)
written in 1557 and published posthumously in 1575 in Venice, 
\cf\nolinebreak\ \cite{maurolycus}. \ 
\pascal\ used structural induction 
for the proofs of his 
``\hskip-.1em\tightemph{Trait\'e du Triangle Arithm\'etique}'' 
written in 1654
and published posthumously in 1665. \ 
While these induction proofs are still presented as
``generalizable examples\closequotecomma
in the demonstration of ``Cons\'equence XII'' we find
---~for the first time in known history~---
a correct verbalization of the related instance of the 
\axiomofstructuralinduction\ \nlbmath{\inpit{\ident{S}}}; \ 
\cf\nolinebreak\ \cite[\p\,103]{pascal},  
\cite[\p\,57]{plato-induction}\@. \ 

\yestop
\subsubsection{Revival}\label{section Revival}%
In the 1650s, \hskip.2em
\pascalname\ \pascallifetime\ (\Paris) 
exchanged letters on probability theory and\emph\descenteinfinie\
with \fermatname\ \fermatlifetime\ (Toulouse), \hskip.2em
who was the first to describe the {\em Method of \DescenteInfinie}\/
explicitly. \
\vietename\ \vietelifetime\ (Paris) had already given 
a new meaning to the word\emph{analysis} 
by extending the analysis of concrete mathematical problems
to the algebraic analysis of the process of their solution. \
\fermat\ improved on \viete:
\par\yestop\noindent
Instead of a set of rules that sometimes did find a single
solution to the ``double
equations'' of \diophantusname\ \diophantuslifetime\ and sometimes did not, 
he \nolinebreak invented a\emph{method}
to enumerate an infinite set of solutions,
which is described in the ``\hskip-.18em\tightemph{Inventum Novum}'' by the 
number theoretician \billyname\ \billylifetime; \ 
for a French translation 
\cfnlb\ \cite[\Vol\,III, \PP{325}{398}]{fermat-oeuvres};
\ for a discussion 
see \cite[\litsectref{VI.III.B}]{fermat-career}\@. 
\pagebreak
\par\yestop\noindent
Much more than that, \fermat\ was the first who
---~instead of just proving a theorem~---
analyzed the {\em method}\/ of proof search. \
This becomes obvious from the description of 
the {\em Method of \DescenteInfinie}\/ in a letter for
\huygensname\ \huygenslifetime\ (Den Haag)
entitled \hskip.2em 
``\nolinebreak\hspace*{-.1em}\nolinebreak
{\em\letterforhuygenstitle}\hskip.2em\closequotesemicolon \hskip.2em
\cfnlb\ \cite[\Vol\,II, \PP{431}{436}]{fermat-oeuvres}\@. \ 
This letter, in which \fermat\ also lists some theorems and claims
to have proved them by {\em\descenteinfinie}, \hskip.1em
was sent to \carcaviname\ in \mbox{August\,\letterforhuygensyear}. \
Moreover, \hskip.1em
in his letter for \huygens\ and in \cite{diophantus-fermat}, \hskip.3em
\fermat\ \nolinebreak was also the first to provide
a correct verbalization of proofs by {\em\descenteinfinie} and to overcome 
their presentation 
as ``generalizable examples\closequotefullstopextraspace
{\em Methodological}\/
considerations seem to have been \fermat's primary concern.

\yestop
\subsection{Linguistic and Refined Logic-Historical Points of View}

\yestop\noindent
The level of abstraction of our previous 
discussion of\emph\descenteinfinie\
is well-suited for the description of the structure of 
mathematical proof search in two-valued logics,
where the difference between a proof by contradiction and a
positive proof of a given theorem is only a linguistic one and
completely disappears when we formalize these proofs in a 
state-of-the-art modern logic calculus, such as the one of 
\cite{wirthcardinal}.
An investigation into the history of mathematics, however, also 
has to consider the linguistic representation and the 
exact logical form of the presentation.

\yestop
\subsubsection{An Inappropriate Refinement by \unguru\ and \acerbi}
\halftop\noindent
Such a linguistic and logic-historical refinement
can easily go over the top. For instance, from the above-mentioned 
fact that we find
---~for the first time in known history~---
a \nolinebreak correct verbalization of the related instance of the 
\axiomofstructuralinduction\ \nlbmath{\inpit{\ident{S}}}
in \pascal's publications, 
it is not sound to conclude that 
\pascal\ was the first to do structural induction 
(as claimed in \cite{unguru-one})
or the first to do it {\em consciously}\/
(as claimed in \cite{plato-induction}). \
These claims are just as abstruse to most working mathematicians as the 
claim that \fermat\ was the first to do {\em\descenteinfinie}. \ 
And \cite{Greeks-induction} is perfectly right to object to this view
on the basis of a deeper understanding of the mathematical activity,
although we 
have to be careful not to interpret modern thinking
into the historical texts.

\yestop\noindent
Mathematics is mostly a top-down procedure and when we do not formalize 
and explicate every bit of it, 
we may have good reasons and be well aware of what we do. \
Human mathematical activity includes subconscious elements,
but this does not mean that their application is unconscious. \
Just as music is not captured by notes
and not necessarily invented as notes,
mathematical activity cannot be captured by its formalization and
is \nolinebreak not necessarily well-expressed in natural
or formal language. \
(Actually, formalization is a dangerous step for a mathematician
 because afterward there is hardly any way back to his original intuition.)
\pagebreak

\yestop
\subsubsection{Our Suggestion for an Unproblematic Classification Scheme}
\halftop\noindent
All in all, for our subject here 
there is actually no need to discuss the working mathematician's consciousness:
It suffices to speak of \begin{enumerate}\noitem\item
quasi-general proofs (\ie\ proofs by generalizable examples), 
\noitem\item general proofs 
(\ie\ proofs we would accept from our students in an examination today), 
\noitem\item
proofs with an explicit statement of the related instance 
of an induction axiom or theorem,
and \noitem\item
proofs with an explicit statement of an induction axiom or theorem itself.
\end{enumerate}

\yestop
\subsubsection{An Appropriate Refinement by \bussottiname}\label
{subsubsection An Appropriate Refinement}
\halftop\noindent
There is evidence that 
such a linguistic and logic-historical refinement
is necessary to understand the fine structure of historical reasoning
in mathematics. \
For instance, in \euclid's Elements,
\litpropref{VIII.7} is just the contrapositive of \litpropref{VIII.6}, 
and this is just one of several cases that we find a proposition with
a proof in the Elements, where today we just see a corollary. \
Moreover, even \fermat\ reported in his letter 
for \huygens\ (\cfnlb\ \sectref{section Revival})
that he had had problems to apply the Method of\emph\DescenteInfinie\
to positive mathematical statements.
\begin{quote}``\frenchtextone''
\getittotheright{\frenchtextonecitation}
\par\halftop\noindent``\frenchtextoneinenglish'' 
\end{quote}
Because of the work of \frege\ and \peano,
these logical differences may be considered trivial today.
Nevertheless, they were not trivial before, and to understand the
history of mathematics and the fine structure in which mathematicians reasoned,
the distinction between affirmative and negative theorems 
and between direct and apagogic methods of demonstration is important.
\pagebreak\par\yestop\noindent
Therefore, it is well justified 
when in \cite{From-Fermat-to-Gauss}, following the above statement of
\fermat,
the Method of\emph\DescenteInfinie\
is subdivided into\emph{indefinite descent} \nlbmath{\inpit{\ident{ID}}} 
and\emph{reduction-descent} \nlbmath{\inpit{\ident{RD}}}:%
\label{page ID}%
\par\halftop\noindent\math{\begin{array}{@{}l@{\ \ \ }r@{\ }l@{}}
  \inpit{\ident{ID}}
 &\forall P\stopq
 &\inparentheses{\headroom
      \forall x\stopq\app P  x
    \nottight{\nottight{\nottight{\nottight\antiimplies}}}
      \exists\tight<\stopq\inparenthesesoplist{
       \forall v\stopq\inparentheses{
            \neg\app P  v
          \nottight{\nottight\implies}
            \exists u\tight<v\stopq\neg\app P  u}
     \oplistund
        \Wellfpp <}}
\\
\\\inpit{\ident{RD}}
 &\forall P\stopq
 &\inparentheses{
      \forall x\stopq\app P  x
    \nottight{\nottight{\nottight{\nottight\antiimplies}}}
      \exists\tight<\stopq\exists S\stopq\inparenthesesoplist{
       \forall u\stopq\inparentheses{
            \app S u\nottight{\implies}\app P u}
     \oplistund
       \forall v\stopq\inparenthesesoplist{
            \neg\app S v\und\neg\app P  v
          \oplistimplies
            \exists u\tight<v\stopq\neg\app P  u}
     \oplistund
        \Wellfpp <}}
\\\end{array}}
\par\yestop\noindent
Actually, ``\Wellfpp<'' does not occur in \cite{From-Fermat-to-Gauss}
because for \fermat\ the Method of\emph\Descenteinfinie\ was actually 
restricted to the \wellfounded\ ordering of the natural numbers.
\par\yestop\noindent
Although \inpit{\ident{N}}, \inpit{\ident{ID}}, and \inpit{\ident{RD}}
are logically equivalent in two-valued logics, 
according to \cite{From-Fermat-to-Gauss}
{\em\descenteinfinie}\/ does not subsume
proofs by \noetherian\ or structural induction. 
This is in opposition to our more coarse-grained discussion above.
With this fine-grained distinction, \hskip.1em
on \litspageref 2 of \cite{From-Fermat-to-Gauss}, \hskip.1em
we find the surprising claim
that there is only a single proof by indefinite descent in the whole
Elements, namely the before-cited Proof of \litpropref{VII.31}\@. \ 
Indeed, at least all those proofs in the Elements beside VII.31 
which I reexamined and which proceed by mathematical induction,
actually proceed by reduction-descent or structural induction,
but not by indefinite descent:
The correctness proofs of the 
\euclid ian Algorithm (\litpropref{VII.2}
and \litpropref{X.3}) are reduction-descents
with a horrible linguistic surface structure.
Similarly, the proofs of \litproprefs{IX.12}{IX.13} are reduction-descents
with superfluous sentences confusing the proof idea.

\yestop\noindent
Note that, as already repeatedly expressed above, a logical formalization 
cannot capture a mathematical method. 
Moreover, 
as also already expressed above,
logical equivalence of formulas does not imply the equivalence of the
formalized methods. 
For an interesting discussion of this difficult subject 
see \cite[\litchapref 7]{From-Fermat-to-Gauss}.

\yestop\noindent
Nevertheless, \inpit{\ident{N}}, \inpit{\ident{ID}}, and \inpit{\ident{RD}}
sketch methods of proof search equivalent for the 
working mathematician of today.
Indeed: \inpit{\ident{ID}}
---~roughly speaking~---
is the \mbox{contrapositive of \inpit{\ident{N}}}, 
which means that in two-valued logics
the methods only differ in verbalization.
Moreover, 
a proof by \inpit{\ident{ID}} is a proof by \inpit{\ident{RD}}
when we set \math S to the empty predicate.
Finally, a proof by \inpit{\ident{RD}} can be transformed into a proof by
\inpit{\ident{ID}} as follows:
Suppose we have proofs for the statements in the conjunction 
of the premise of \nlbmath{\inpit{\ident{RD}}}.
The proofs of 
\ \mbox{\math{\forall u\stopq\inparentheses{
\app S u\nottight{\implies}\app P u}}} \ 
\\and \hfill{\maths{\forall v\stopq\inparenthesesoplist{
            \neg\app S v\und\neg\app P  v
          \oplistimplies
            \exists u\tight<v\stopq\neg\app P  u}}{} \hfill
give a proof of \hfill\maths{\forall v\stopq\inparenthesesoplist{
            \neg\app S v\und\neg\app P  v
          \oplistimplies
            \exists u\tight<v\stopq\inpit{\neg\app S v\und\neg\app P  u}}}.}\\
Instantiating the \math P in \inpit{\ident{ID}} via
\bigmaths{\{P \mapsto\lambda z\stopq\inpit{\app S z\oder\app P z}\}},
the latter proof can be schematically transformed into a proof of \bigmaths
{\forall x\stopq\inpit{\app S x\oder\app P x}}{} by 
\bigmaths{\inpit{\ident{ID}}}. \ 
And then from the proof of \ \mbox{\math{\forall u\stopq\inparentheses{
\app S u\nottight{\implies}\app P u}}} \ again,
we get a proof of \bigmaths{\forall x\stopq\app P  x}, as intended. \ 
Thus, in any case, the resulting proof does not significantly 
differ in the mathematical structure from the original one.
\pagebreak\par\yestop\noindent
Note that this is contrary to the case of \noetherian\ \vs\ structural 
induction,
where the only transformation I see from the former to the latter 
\ (the other direction is trivial, 
   \cfnlb\ \cite{wirthcardinal}, \litsectref{1.1.3}) \ 
is to show that the\emph{axiom} \inpit{\ident{S}} implies \Wellfpp\ssymbol,
and then leave the application of \inpit{\ident{N}} unchanged.
This 
transformation, however, is not complete because it does not remove the
application of \inpit{\ident{N}}, which is a\emph{theorem} anyway.

\yestop\noindent
All in all, this shows that
---~while structural and \noetherian\ induction
vastly differ in practical applicability~---
for a working mathematician today
it is not important for his proof search to be aware of
the differences between 
\noetherian\ induction \nlbmath{\inpit{\ident{N}}}, 
indefinite descent \nlbmath{\inpit{\ident{ID}}}, and 
reduction-descent \nlbmath{\inpit{\ident{RD}}}. \ 
And thus, we will continue to subsume all the three
under the Method of\emph\DescenteInfinie.

\yestop
\subsubsection{A Further Refinement}\label
{subsection A Further Refinement}
\halftop\noindent
For the soundness of \walsh's interpretation of \fermat's proof in
our \sectref{section Interpretation}, \ 
we have to invent the following further refinement
to the logic-historical discussion of\emph\descenteinfinie.
\par\yestop\noindent
The predicate \nlbmath P
in the theorem \nlbmath{\inpit{\ident{ID}}} 
of \nolinebreak\sectref{subsubsection An Appropriate Refinement}
may actually vary in the indefinite
descent, in \nolinebreak the sense that
---~for a function \nlbmath P from natural numbers to predicates~---
we have the following theorem:
\par\noindent\math{\begin{array}{@{}l@{\ \ \ }r@{\ }l@{}}
  \inpit{\ident{ID}'}~~~
 &\forall P\stopq
 &\inparenthesesoplist{\headroom
      \forall x\stopq\app{P_0}x
    \oplistantiimplies
      \exists\tight<\stopq\inparenthesesoplist{
       \forall i\tightin\N\stopq\forall v\stopq\inparentheses{
             \neg\app{P_{i}}v
          \nottight{\nottight\implies}
            \exists u\tight<v\stopq\neg\app{P_{i+1}}u}
     \oplistund
        \Wellfpp <}}
\\\end{array}} 
\par\halftop\noindent
For a sufficiently expressible logic, this again makes no difference: \ 
Indeed, to prove theorem \nlbmath{\inpit{\ident{ID}'}} and even to use
theorem \nlbmath{\inpit{\ident{ID}}} instead of it
without a significant change of the structure of the proof,
it suffices to instantiate theorem \nlbmath{\inpit{\ident{ID}}} 
according to 
\par\noindent\LINEmaths{\displayset
{P\nottight{\nottight{\nottight\mapsto}}\lambda x\stopq
\forall i\tightin\N\stopq
\app{P_i}x}}.\par\noindent
Typically
---~as in the example of \sectref{section Interpretation}~---
the set \nlbmath{\setwith{P_i}{i\tightin N}} is finite. \ 
In this case, the universal quantification in 
\bigmaths{\forall i\tightin\N\stopq\app{P_i}x}{}
can be replaced with a finite conjunction.
\vfill\pagebreak
\yestop\yestop\yestop\section{Prerequisites from Number Theory}\label
{section Prerequisites from Number Theory}
\halftop\noindent
In this \sectref{section Prerequisites from Number Theory}, \hskip.2em
we list the propositions and the proofs that I found in my 
easy chair without any help beside \euclid's Elements. 
\par\yestop\noindent
If the reader is experienced in number theory or wants to 
have the pleasure of doing 
some exercises in elementary number theory on his own, 
he should skip this 
\sectref{section Prerequisites from Number Theory}
and continue directly with \nolinebreak\sectref{section fermat's Proof}. \
Moreover, we generally recommend to skip this 
\sectref{section Prerequisites from Number Theory}
on a first reading.
\par\yestop\noindent
We follow the Elements quite closely, 
but occasionally deviated from them if an
alternative course is more efficient. \hskip.3em
As we address modern readers, 
the language of our presentation,
however,  
is a modern one, 
most atypical for the \nth{17}\,century. \
Furthermore, 
in this \nlbsectref{section Prerequisites from Number Theory}, \hskip.2em
we \nolinebreak follow the historical proof ideas only very roughly and 
are not seriously concerned with historical authenticity. \
Nevertheless, \hskip.1em
as expressed already in \sectref{section Introduction}, \hskip.15em
we hope that our reconstruction of elementary number theory 
in this \nlbsectref{section Prerequisites from Number Theory} \hskip.1em
does not essentially differ from what
\fermat's contemporary mathematicians could have
achieved if \fermat\ had been able to interest 
them in his new number theory. \ 
This aspect becomes crucial, however, only for 
\fermat's proof in \sectref{section fermat's Proof}.
\par\yestop\noindent
The proofs missing here can be found \sectref{section missing proofs} of
the appendix.


\yestop\subsection{From the Elements, \Vol\,VII}

\halftop\halftop\noindent
Let all variables range over the set of natural numbers \N\ 
(including \nlbmath 0), unless indicated otherwise. \ 
Let `\math\prec' denote the (irreflexive) ordering 
and `\math\preceq' the reflexive ordering on \nolinebreak\N\@. \  
Let \bigmaths{\posN:=\setwith{n\tightin\N}{0\tightnotequal n}}.

\halftop\halftop
\begin{definition}[Divides]\\
{\em\math x divides y}\/ \ (written: \bigmaths{x\mid y})
\udiff\ there is a \math k \hskip.1em
such that \bigmaths{k x\tightequal y}.
\end{definition}

\begin{corollary}\label{corollary divides} \
The binary relation \math\mid\ is a reflexive ordering on the natural numbers
with minimum \math 1 and maximum \math 0. \
Moreover, \bigmaths{\inpit{x\mid y}\und\inpit{y\tightnotequal 0}}{} implies 
\bigmaths{x\tightpreceq y}.
\end{corollary}

\begin{corollary}\label{corollary trivial one} \
Let us assume \bigmaths{x\mid y_0}. \
Then \ \bigmaths{x\mid y_1}{} \uiff\ \bigmaths{x\mid y_0\tight+ y_1}.
\end{corollary} 

\begin{corollary}\label{corollary trivial two}
\bigmaths{\inpit{x\tightequal 0}\oder\inpit{y\mid z}}{}
\uiff\ \bigmaths{x y\mid x z}.
\end{corollary} 

\halftop
\begin{definition}[Coprime]\\
\bigmaths{l_1,\ldots,l_n}{} are\emph{coprime} \udiff\
\bigmaths{\forall x\stopq\inparentheses{\forall i\tightin\{1,\ldots,n\}\stopq
\inpit{x\mid l_i}\nottight{\nottight{\nottight{\nottight{\implies}}}}
x\tightequal 1}}.\end{definition}

\begin{corollary}\label{corollary trivial three}
If \math{p,q} are coprime and
 \bigmaths{p\tightsucceq q}, then \bigmaths{p\tightsucc q}{} or
\bigmaths{p\tightequal q\tightequal 1}.\end{corollary}

\pagebreak
\begin{lemma}[\euclid's Elements, \litproprefs{VII.20}{VII.21}, generalized]
\label{lemma Euclid VII 20}\label{lemma Euclid VII 21}
\par\noindent Let \bigmaths{x_0 y_1\tightequal y_0 x_1}{} with 
\math{x_0,x_1,y_0,y_1\in\posN}.
\par\noindent Then the following holds:
\begin{itemize}\noitem\item[(i)~]For every \math{z_0,z_1}, \hskip.2em
we have: 
\ \bigmaths{y_0 z_1\tightequal z_0 y_1}{} \uiff\ 
\bigmaths{x_0 z_1\tightequal z_0 x_1}.\noitem\end{itemize}
\par\noindent Moreover, 
the following two cases are equivalent:
\begin{itemize}\noitem\item[(ii)~]\math{y_0,y_1} are coprime.
\noitem\item[(iii)~]
\math{y_0} is\/ \tight\preceq-minimal 
such that there is a\/ \math{y_1'\in\posN} with
\bigmaths{x_0 y_1'\tightequal y_0 x_1}.\noitem\end{itemize}
Furthermore, 
in each of the two cases (ii) and (iii), the following holds:
\begin{itemize}\noitem\item[(iv)~]
There is a\/ \math{k\in\posN} with \bigmaths{k y_i\tightequal x_i}{}
for\/ \math{i\tightin\{0,1\}}.
\end{itemize}\end{lemma}


\halftop
\begin{lemma}[\euclid's Elements, \litpropref{VII.23}]\label
{lemma Euclid VII 23}\\
If\/ \math{y,z} are coprime and \bigmaths{x\mid y},
then \math{x,z} are coprime,
too.\end{lemma}

\halftop
\begin{lemma}[\euclid's Elements, \litpropref{VII.24}, generalized]\label
{lemma Euclid VII 24}\\
If\/ \math{x_i,z} are coprime for\/ \math{i\in\{1,\ldots,m\}}, \ 
then\/ \bigmaths{\prod_{i=1}^m x_i\comma z}{} are coprime, too.\end{lemma}


\halftop
\begin{lemma}[\euclid's Elements, \litpropref{VII.26}, generalized]\label
{lemma Euclid VII 26}\\
If\/ \math{x_i,y_j} are coprime for\/ \math{i\in\{1,\ldots,m\}}{}
and\/ \math{j\in\{1,\ldots,n\}},
then \bigmaths{\prod_{i=1}^m x_i\comma\prod_{i=1}^n y_i}{}
are coprime, too.\end{lemma}

\begin{corollary}[\euclid's Elements, \litpropref{VII.27}]\label
{corollary Euclid VII 27}\\
If\/ \math{y,z} are coprime, then\/ \math{y^{n},z^{m}} are coprime, 
too.\end{corollary}

\halftop
\begin{definition}[Prime]\\
A number \math p is\emph{prime} \udiff\
\bigmaths{p\tightnotequal 1}{} and
\ \bigmaths{\inpit{x\mid p}{{{\nottight{\implies}}}}
x\tightin\{1,p\}}{} for all \nlbmath x.\end{definition}

\begin{corollary} \
If\/ \math p is prime, \hskip.2em
then \bigmaths{2\tightpreceq p}.
\end{corollary}

\halftop
\begin{lemma}[\euclid's Elements, \litpropref{VII.29}]\label
{lemma Euclid VII 29}\\
If\/ \math p is prime and \bigmaths{p\nmid x}, then
\math{p,x} are coprime.\end{lemma}

\halftop\noindent
The following lemma is popular today under the label of 
``{\it\euclid's Lemma}\closequotecolon

\begin{lemma}[\euclid's Elements, \litpropref{VII.30}]\label
{lemma Euclid VII 30}\\
If \math p is prime and \bigmaths{p\mid x_1 x_2}, then
\bigmaths{p\mid x_1}{} or \bigmaths{p\mid x_2}.
\end{lemma}

\halftop\noindent The following lemma will be applied 
exclusively in the proofs of 
\lemmrefss{lemma divides}{lemma coprime}{lemma mahoney footnote one two}.
\begin{lemma}[\euclid's Elements, \litpropref{VII.31}]\label
{lemma Euclid VII 31}\\
For any \math{x\tightnotequal 1}, there is some prime \math p
such that \bigmaths{p\mid x}.\end{lemma}

\yestop\yestop\yestop\yestop\subsection{From the Elements, \Vol\,VIII}
\yestop\begin{definition}[Continued Proportion]\\
\math{x_0,\ldots,x_{n+1}} are in\emph{continued proportion}
\udiff\ \bigmaths
{\forall i\tightin\{0,\ldots,n\tight+1\}\stopq\inpit{x_i\tightin\posN}}{}
and\\
\bigmaths{\forall i\tightin\{1,\ldots,n\}\stopq
\inpit{x_{i-1}x_{i+1}\tightequal x_i^2}}.
\end{definition}

\yestop\yestop\noindent The following lemma will be applied exclusively in
the proof of \lemmref{lemma Euclid VIII 2}.
\begin{lemma}[\euclid's Elements, \litpropref{VIII.1}, generalized]\label
{lemma Euclid VIII 1}\\
If\/ \math{x_0,\ldots,x_{n+1}} and \math{y_0,\ldots,y_{n+1}} are in
continued proportion, if \bigmaths{x_0 y_1\tightequal y_0 x_1}, 
and if\/ \math{x_0,x_{n+1}} are coprime, then there is some \math{k\in\posN}
with \bigmaths{k x_i\tightequal y_i}{} 
for\/ \math{i\in\{0,\ldots,n\tight+1\}}.\end{lemma}

\yestop\yestop\noindent 
The following lemma will be applied exclusively in
the proof of \lemmref{lemma additional one}.
\begin{lemma}[\euclid's Elements, \litpropref{VIII.2}, generalized]\label
{lemma Euclid VIII 2}\\
If\/ \math{x_0,\ldots,x_{n+1}} are in continued proportion, 
then there are \math{k\in\posN} and coprime
\math{y,z\in\posN} such that
\math{y^{n+1} z^0, \ldots, y^{n+1-i} z^{i},\ldots,y^0 z^{n+1}}
are in continued proportion, too,
\bigmaths{x_0\inpit{y^n z^1}\tightequal\inpit{y^{n+1} z^0}x_1},
\bigmaths{k y^{n+1-i} z^{i}\tightequal x_i}{} 
for\/ \math{i\in\{0,\ldots,n\tight+1\}},
and, moreover, in case that \math{x_0,x_{n+1}} are coprime,
\bigmaths{k\tightequal 1}.\end{lemma}

%

\yestop
\begin{lemma}[\euclid's Elements, \litpropref{VIII.14}]\label
{lemma Euclid VIII 14}\\
\bigmaths{x\mid y}{} \uiff\ \bigmaths{x^2\mid y^2}.
\end{lemma}

\yestop
\subsection{Further Simple Lemmas for \fermat's Proof}

\yestop\begin{lemma}\label{lemma additional one}\ \ 
If\/ \math{a,b} are coprime and \bigmaths{a b\tightequal x^2},
then there are coprime \math{y,z} with \bigmaths{a\tightequal y^2},
\bigmaths{b\tightequal z^2},
and \bigmaths{x\tightequal y z}.\end{lemma}

\yestop\begin{lemma}\label{lemma coprime} \ 
\bigmaths{l_1,\ldots,l_n}{} are not coprime \uiff\
\bigmaths{\exists p\mbox{ prime}\stopq\forall i\tightin\{1,\ldots,n\}\stopq
\inpit{p\mid l_i}}.\end{lemma}

\yestop\begin{lemma}\label{lemma coprime two} \ 
Let \maths{a,b}{} be coprime. Let \math p be a prime number.\begin{enumerate}
\noitem\item[(1)]
Either \math{p a,b} are coprime or \math{a, p b} are coprime.
\item[(2)]
If \bigmath{p a b\tightequal v^2} for some \nlbmath v, \ 
then there are\/ \math{m\in\N} and\/ \math{k\in\posN} 
\\such that\/ 
\math{p m,k} are coprime and \bigmaths{\{p m^2,k^2\}=\{a,b\}}.
\end{enumerate}\end{lemma}

\yestop\begin{lemma}\label{lemma mahoney footnote one two}
Suppose \bigmaths{a\tightsucceq b},
\bigmaths{x\mid a\tight-b}, and \bigmaths{x\mid a\tight+b}. Then we have:
\begin{enumerate}\noitem\item[(1)]
\bigmaths{x\mid 2a}{} and 
\bigmaths{x\mid 2b}.
\item[(2)] \ 
If \math{a,b} are coprime, then
\bigmaths{x\tightpreceq 2}.
\halftop\end{enumerate}\end{lemma}

\yestop\begin{lemma}\label{lemma p and q coprime}\\
If\/ \math p, \math q are coprime with \bigmaths{p\tightsucceq q}, 
then\/ \math{p q}, \math{p^2\tight-q^2} are coprime, 
and\/ \math{p q}, \math{p^2\tight+q^2} are coprime.\end{lemma}

\yestop
\subsection{Roots of \ \math{x_0^2+x_1^2\protect\tightequal x_2^2}}\label
{section roots one}%
\yestop\begin{lemma}\label{lemma even one}
\bigmaths{x_0^2+x_1^2\tightequal x_2^2}{} \uiff\ 
for some \math{i\tightin\{0,1\}},
there are \math{a,b} such that \bigmaths{a\tightsucceq b},
\ \mbox{\maths{x_i\tightequal 2\sqrt{a b}},} \ 
\bigmaths{x_{1-i}\tightequal a\tight-b},
and 
\bigmaths{x_2\tightequal a\tight+b}.
\end{lemma}
\begin{proofqed}{\lemmref{lemma even one}} \ 
The ``if''-direction is trivial.
Let us assume \bigmaths{x_0^2+x_1^2\tightequal x_2^2}{} 
to show the ``only if''-direction.
Suppose \bigmaths{x_i\tightequal 2y_i\tight+1}{} for \math{i\tightin\{0,1\}}.
Then \bigmaths{x_0^2+x_1^2
=\inpit{2y_0\tight+1}^2+\inpit{2y_1\tight+1}^2
=4y_0^2\tight+4y_0\tight+1
+4y_1^2\tight+4y_1\tight+1
=2\inpit{2\inpit{y_0^2\tight+y_0\tight+y_1^2\tight+y_1}+1}
}. As the square of an even number divides by \math 4 and the 
square of an odd number is odd, this would mean that \bigmaths{x_0^2+x_1^2}{}
is not a square.
Thus, there is some \math{i\in\{0,1\}} and some \math c with
\bigmaths{x_i\tightequal2c}.
But then \bigmaths{x_2\tight\pm x_{1-i}}{} must be even too,
because \bigmaths{\inpit{x_2\tight+x_{1-i}}\inpit{x_2\tight-x_{1-i}}
=x_2^2-x_{1-i}^2=x_i^2=4c^2}{} means that one of them must be even
by \lemmref{lemma Euclid VII 30},
and then the other is even, too.
Thus, there are \math{a,b} such that 
\bigmaths{x_2\tight+x_{1-i}\tightequal 2a}{} and 
\bigmaths{x_2\tight-x_{1-i}\tightequal 2b}. 
This implies 
\bigmaths{2x_2\tightequal 2a\tight+2b},
\bigmaths{2x_{1-i}\tightequal 2a\tight-2b}, and
\bigmaths{2a2b\tightequal 4c^2}, and then
\bigmaths{x_2\tightequal a\tight+b},
\bigmaths{x_{1-i}\tightequal a\tight-b}, and
\bigmaths{a b\tightequal c^2}.
\end{proofqed}

\yestop\yestop\noindent
Note that in \lemmref{lemma even one}
we cannot require any of \math a and \math b to be a square in general.
For instance, for 
\bigmaths{x_0\tightequal 12\und x_1\tightequal 9\und x_2\tightequal 15}, 
we have \bigmaths{x_0^2+x_1^2\tightequal x_2^2},
but necessarily get \ \mbox{\maths{\sqrt{a b}\tightequal 6}{}} \ 
(as \math{x_1} is odd), 
and then, if any of \math a or \math b is a square, we have 
\bigmaths{
\inpit{a\tightequal  9\und b\tightequal 4}\oder
\inpit{a\tightequal 36\und b\tightequal 1}},
\ie\ 
\bigmaths{
\inpit{x_2\tightequal 13\und x_1\tightequal 5}\oder
\inpit{x_2\tightequal 37\und x_1\tightequal 35}},
which are \pythagorean\ triangles not similar to 
the original \bigmaths{x_2\tightequal 15\und x_1\tightequal 9}. \ 
But \bigmaths{a\tightequal 12\und b\tightequal 3}{} provide the 
generators for 
\bigmaths{x_0\tightequal 12\und x_1\tightequal 9\und x_2\tightequal 15}, 
whose existence is guaranteed by \lemmref{lemma even one}.

\yestop\noindent
If \math{x_0,x_1,x_2} are coprime, however,
then \math{a,b} must be coprime 
\ \ (as \ 
 \math{\inpit{y\mid a}\nottight{\und}\inpit{y\mid b}
 \ \ \implies}\\\math{
 \forall i\tightin\{0,1,2\}\stopq\inpit{y\mid x_i}}) \ \ 
and one even and one odd
\ (as otherwise \math{\forall i\tightin\{0,1,2\}\stopq\inpit{2\mid x_i}})\@. \ 
And then they must be squares because of
\ \mbox{\math{x_i\tightequal 2\sqrt{a b}}} \ and  
\lemmref{lemma additional one}; \ 
say \bigmaths{a\tightequal p^2}{}
and \ \mbox{\math{b\tightequal q^2}}. \ 
Then \math{p,q} are coprime and one even and one odd, too.
All in all, we get as a corollary of \lemmref{lemma even one}:
\begin{corollary}\label{corollary even two two}
If \bigmaths{x_0^2+x_1^2\tightequal x_2^2}{} 
and \math{x_0,x_1,x_2} are coprime, then, for some \math{i\tightin\{0,1\}},
there are coprime \math{p,q} such that
one of them is odd and one of them is even, \bigmaths{p\tightsucc q},
\ \mbox{\maths{x_i\tightequal 2p q},} \ 
\bigmaths{x_{1-i}\tightequal p^2\tight-q^2},
and \bigmaths{x_2\tightequal p^2\tight+q^2}.\end{corollary}
Note that in \cororef{corollary even two two} we cannot require
\bigmaths{q\tightin\posN}{} because for the case of
\bigmaths{\trip{x_0}{x_1}{x_2}\tightequal\trip 1 0 1}{}
we have \bigmaths{x_0^2+x_1^2\tightequal x_2^2}{} 
and \math{x_0,x_1,x_2} are coprime, \ 
but \bigmaths{i\tightin\{0,1\}}, \bigmaths{p\tightsucc q},
\ \mbox{\maths{x_i\tightequal 2p q},} \ 
\bigmaths{x_{1-i}\tightequal p^2\tight-q^2},
and \bigmaths{x_2\tightequal p^2\tight+q^2}{} implies
\bigmaths{q\tightequal 0}.

\yestop\begin{lemma}[{\cite[\litpropref{XXXVIII}]{frenicle}}, generalized]\label
{lemma frenicle XXXVIII}
\\\mediumheadroom Let \bigmaths{x_0^2+x_1^2\tightequal x_2^2}{} 
and \math{x_0,x_1,x_2} be coprime.
Then exactly one of \nlbmath{x_0,x_1} is even, say \math{x_i}.
\\\mediumheadroom If \bigmaths{x_i\tightequal v^2}{} for some \nlbmath v, \ 
then there are \math{m\in\N} and \math{k\in\posN}
such that\/ \math{2m,k} are coprime,
\bigmaths{\inpit{m\tightequal 0}\equivalent\inpit{x_0\tightequal 0}}, and\/
\bigmaths{\headroom\inpit{2m^2}^2\tight+\inpit{k^2}^2\tightequal x_2}.
\end{lemma}

\begin{proofqed}{\lemmref{lemma frenicle XXXVIII}} \ 
By \cororef{corollary even two two}, 
there are \math{j\in\{0,1\}} and 
coprime \math{p,q} such that
one of them is odd and one of them is even, \bigmaths{p\tightsucc q},
\ \mbox{\maths{x_j\tightequal 2p q},} \ 
\bigmaths{x_{1-j}\tightequal p^2\tight-q^2},
and \ \mbox{\math{x_2\tightequal p^2\tight+q^2}}. \ 
We have assumed \math{x_i} to be the even one in
\math{\{x_0,x_1\}}, \ 
\ie\ \bigmaths{i\tightequal j}.
Thus, \ \mbox{\math{2p q\tightequal v^2}} \ by assumption. \ 
By \lemmref{lemma coprime two}(2),
there are \math{m\in\N} and \math{k\in\posN} such that 
\math{2 m,k} are coprime and \bigmaths{\{2 m^2,k^2\}=\{p,q\}}.
Moreover the following are logically equivalent:
\bigmaths{x_0\tightequal 0},
\bigmaths{q\tightequal 0},
\ \mbox{\math{m\tightequal 0}}. \ \ 
Finally, 
\bigmaths{\mediumheadroom\inpit{2m^2}^2\tight+\inpit{k^2}^2\tightequal 
p^2\tight+q^2\tightequal x_2}.
\end{proofqed}

\yestop\yestop\yestop
\subsection{Roots of \ \math{x_0^2+2x_1^2\protect\tightequal x_2^2}}
\yestop\yestop\begin{lemma}\label{lemma fermat help two}\\
\bigmaths{x_0^2+2x_1^2\tightequal x_2^2}{} \uiff\ 
there are \math{a,b} such that \bigmaths{a\tightsucceq b},
\ \mbox{\math{x_0\tightequal a\tight-b}}, \ 
\bigmaths{x_1\tightequal \sqrt{2 a b}},
and 
\bigmaths{x_2\tightequal a\tight+b}.
\end{lemma}
\begin{proofqed}{\lemmref{lemma fermat help two}} \ 
The ``if''-direction is trivial.
Let us assume \bigmaths{x_0^2+2x_1^2\tightequal x_2^2}{} to show the 
``only if''-direction.
\bigmaths{\inpit{x_2\tight+x_{0}}\inpit{x_2\tight-x_{0}}
\tightequal x_2^2-x_{0}^2\tightequal 2x_1^2}{} 
means that one of \math{x_2\tight\pm x_{0}}
must be even by \lemmref{lemma Euclid VII 30},
and then the other is even, too.
Thus, there are \math{a,b} with
\bigmaths{x_2\tight+x_{0}\tightequal 2 a},
and 
\ \mbox{\maths{x_2\tight-x_{0}\tightequal 2b}.} \ \ 
But then 
\bigmaths{2 x_2\tightequal 2 a\tight+2 b},
\bigmaths{2 x_0\tightequal 2 a\tight-2 b}, and
\bigmaths{2 a 2 b\tightequal 2 x_1^2},
\ie\ 
\ \mbox{\math{x_2\tightequal a\tight+b}}, \ 
\ \mbox{\maths{x_0\tightequal a\tight-b},} \  and
\bigmaths{2 a b\tightequal x_1^2}.
\end{proofqed}

\yestop\yestop\yestop\yestop\noindent
If \math{x_0,x_1,x_2} are coprime, however,
then \math{a,b} must be coprime.
By \lemmref{lemma coprime two}(2),
then there are \math{m\in\N} and \math{k\in\posN} such that 
\math{2 m,k} are coprime and \bigmaths{\{2 m^2,k^2\}=\{a,b\}}.
All in all, we get as a corollary of \lemmref{lemma fermat help two}:
\begin{corollary}\label{corollary fermat help four} \ 
If \bigmaths{x_0^2+2x_1^2\tightequal x_2^2}{} 
and\/ \math{x_0,x_1,x_2} are coprime, 
then there are\/ \math{m\in\N} and\/ \math{k\in\posN} \ 
such that \
\math{2m,k} are coprime,
\bigmaths{2m^2\tightnotequal k^2},
\ \mbox{\math{x_0\tightequal\CARD{2m^2\tight-k^2}}}, \ 
\bigmaths{x_1\tightequal 2 m k},
and 
\ \mbox{\math{x_2\tightequal 2m^2\tight+k^2}}.\end{corollary}
Note that in \cororef{corollary fermat help four} we cannot require
\bigmaths{m\tightin\posN}{} because for the case of
\bigmaths{\trip{x_0}{x_1}{x_2}\tightequal\trip 1 0 1}{}
we have \bigmaths{x_0^2+2x_1^2\tightequal x_2^2}{} 
and \math{x_0,x_1,x_2} are coprime, \ 
but \bigmaths{x_2\tightequal 2m^2\tight+k^2}{} implies
\ \mbox{\math{m\tightequal 0}}.\vfill\cleardoublepage

\section{\fermat's Proof}\label
{section fermat's Proof}

\yestop\noindent
In this \sectref{section fermat's Proof}, \hskip.2em
we first state \fermat's theorem 
of \OBS\ in \cite{diophantus-fermat}
in \nolinebreak modern notation (\sectref{section theorem}). \
Then, we present \fermat's original short French announcement of the theorem
(and the idea of proving it by {\it\descenteinfinie}\/), \hskip.2em
and translate it into English (\sectref{section French statement}). \

In \sectref{section original proof}, \hskip.2em
we present \fermat's original Latin proof. \

As this proof is hard to understand,
we first have to grasp the mathematical ideas implicitly expressed 
in this proof. \
Therefore, \hskip.1em
in \nlbsectref{subsection A Simple Self-Contained Modern Proof}, \hskip.2em
we continue with a simple, 
self-contained, modern English proof of the theorem. \
Afterward, \hskip.1em
in \nlbsectref{subsection An Annotated Translation}, \hskip.2em
we present our translation of the Latin proof,
annotated with 
our interpretation, which is more or less standard.

Note that the cognitive process behind this annotation 
seems to be similar to the interpretation of a 
music passage from its notes in the following sense: \
If we perceive a gestalt
of the passage, this gestalt will be meaningful, but not necessarily the 
original one of the author. \
After projecting our image onto the original passage,
we can then evaluate its adequacy. \
Therefore, \hskip.1em
we look at interpretations of \fermat's proofs in the 
literature and discuss similar interpretations in 
\nlbsectref{section Similar Interpretations of the Proof in the Literature}, 
\hskip.2em
and \walsh's alternative interpretation in 
\nlbsectref{section Interpretation}. \

Finally, \sectref{section fermat's Proof} closes with \frenicle's
more elegant proof of the theorem in \nlbsectref{section aside}.

\yestop\yestop
\subsection{\fermat's Theorem of \OBS\ in \protect\cite{diophantus-fermat}}\label
{section theorem}%

\yestop\noindent
\fermat's theorem 
simply says that
the area of a \pythagorean\ triangle with positive integer side lengths 
is not the square of an integer, or in modern formulation:

\yestop
\begin{theorem}
\label{theorem Fermat one} \
If \bigmaths{x_0,x_1\tightin\posN}{} and
\bigmaths{x_2,x_3\tightin\N}{} and
\bigmaths{x_0^2+x_1^2\tightequal x_2^2}, then
\bigmaths{x_0 x_1\tightnotequal 2x_3^2}.
\end{theorem}
\yestop\noindent
Note that we cannot generalize \theoref{theorem Fermat one} 
by admitting \bigmaths{0\tightin\{x_0,x_1\}}{}
because we have
\par\noindent\LINEnomath{\bigmaths{\quar{x_0}{x_1}{x_2}{x_3}\in\{\quar 0 0 0 0,
\quar 0 1 1 0, \quar 1 0 1 0\}}{} \ \uiff\ \ \bigmaths{
\inpit{x_0^2+x_1^2\tightequal x_2^2}\und\inpit{x_0 x_1\tightequal 2x_3^2}}.}

\yestop\yestop
\subsection{French Abstract of the Theorem and its Proof Idea}\label
{section French statement}%

\yestop\noindent
\fermat\ summarized his original proof 
\hskip.1em
(which we will quote in \nlbsectref{section original proof}) \hskip.2em
in his letter for \huygens\ 
\hskip.1em (which we have already discussed in \nlbsectref{section Revival}):
\halftop\begin{quote}``\frenchtexttwo''\getittotheright
{\letterforhuygensproofcitation}
\pagebreak
\par\halftop\noindent``\frenchtexttwoinenglish''
\end{quote}

\yestop
\subsection{\fermat's Original Latin Proof}\label{section original proof}

\yestop\noindent
\fermat\ wrote in \OBSlong\ of his\emph{Observations on \diophantus}; \ 
 \cfnlb\ \OBSciteinsclose\ \hskip.2em
 (there is a manipulated facsimile in \cite[\p\,78]{weil-number-theory}
 and a true one in \cite[\p\,60]{catherine-goldstein}) \hskip.2em 
 and \hskip.1em
 \OBScitzwei:
\yestop\begin{quote}
{\em\latintextone}\par{\em\latintexttwo}\par{\em\latintextthree}
\yestop\end{quote}
Note that the separation into paragraphs is not original, but intended to 
simplify the 
comparison with the English translation in 
\sectref{subsection An Annotated Translation}, \hskip.2em
which follows the same separation into paragraphs. \
Moreover, \hskip.1em
we have omitted the beginning and the end of \OBS, \hskip.1em
which state the theorem and that it is proved indeed, \hskip.1em
respectively.
\vfill\pagebreak

\subsection{A Simple Self-Contained Modern Proof 
of \theoref{theorem Fermat one}}\label
{subsection A Simple Self-Contained Modern Proof}%

We show \theoref{theorem Fermat one} by {\em\descenteinfinie},
more precisely by indefinite descent (ID, \cfnlb\ \p\pageref{page ID}).

\noindent
Assuming the existence of \math{x_0,x_1,x_2,x_3}
with \ \math{x_0,x_1\in\posN} \ and
\ \mbox{\maths{x_0^2+x_1^2\tightequal x_2^2}{}} \  and \ 
\mbox{\math{x_0 x_1\tightequal 2x_3^2}}, \  
we will show the existence of \math{y_0,y_1,y_2,y_3}
with \ \math{y_0,y_1\in\posN} \ and 
\ \mbox{\maths{y_0^2+y_1^2\tightequal y_2^2},} \ 
\bigmaths{y_0 y_1\tightequal 2y_3^2}, 
and \bigmaths{y_2\tightprec x_2}.

\noindent
First, let us consider the case that there is some prime number \nlbmath z
that divides \math{x_0,x_1}, \ie\ that there are \math{y_i} 
with \bigmaths{x_i\tightequal z y_i}{} for \math{i\tightin\{0,1\}}\@. \ 
Then we have \bigmaths{z^2\inpit{y_0^2+y_1^2}\tightequal x_2^2},
\ie\ \bigmaths{z^2\mid x_2^2}. By \lemmref{lemma Euclid VIII 14},
we get \bigmaths{z\mid x_2}. Thus, there is some \math{y_2\in\posN}
with \bigmaths{x_2\tightequal z y_2}. Then we also have 
\bigmaths{z^2\inpit{y_0^2+y_1^2}\tightequal z^2y_2^2},
\ie\ \bigmaths{y_0^2+y_1^2\tightequal y_2^2}.
Moreover, we have 
\bigmaths{z^2 y_0 y_1\tightequal 2x_3^2}, \ie\ 
\bigmaths{z^2\mid 2 x_3^2}.
As \math z \nolinebreak is prime, from the latter we get
\bigmaths{z\mid 2}{} or \bigmaths{z\mid x_3^2}{} 
by \lemmref{lemma Euclid VII 30}. \ 
By \cororef{corollary trivial two},
\bigmaths{z\tightequal 2}{} and \bigmaths{z^2\mid 2 x_3^2}{} implies
\bigmaths{z\mid x_3^2}. 
Thus, we have \bigmaths{z\mid x_3^2}{} in both cases, and then
\bigmaths{z\mid x_3}{} by \lemmref{lemma Euclid VII 30} again. \ 
Thus, there is some \math{y_3}
with \ \mbox{\math{x_3\tightequal z y_3}}\@. \ \ 
Then \bigmaths{z^2 y_0 y_1\tightequal 2x_3^2\tightequal z^2 2y_3^2},
\ie\ \bigmaths{y_0 y_1\tightequal 2y_3^2}. 
From \bigmaths{x_0,x_1\tightin\posN},
we get \bigmaths{x_2\tightin\posN}.
For each \math{i\tightin\{0,1,2\}}, \
we get \bigmaths{y_i\tightin\posN}{}
from \bigmaths{x_i\tightin\posN}.
Finally, we have \bigmaths{y_2\tightprec x_2}{}
because of \bigmaths{x_2\tightequal z y_2}.
This completes this case by {\it\descenteinfinie}.

\noindent
Thus, we may assume \math{x_0,x_1} to be coprime by \lemmref{lemma coprime},
and
---~\afortiori~--- 
\math{x_0,x_1,x_2} to be coprime, too.

\initial{\underline{Claim\,I:}} 
There are coprime \math{p,q} 
such that
one of them is odd and one of them is even,
\bigmaths{p\tightsucc q},
and there are some \math{c,e,f} 
with
\bigmathnlb{x_2\tightsucc e\tightsucc f\tightsucc 0}{} such that
\par\noindent\LINEnomath{
\bigmaths{p\tightequal e^2},
\bigmaths{q\tightequal f^2}, and 
\bigmaths{p^2\tight-q^2\tightequal c^2}.}
\initial{\underline{Proof of Claim\,I:}}
By \cororef{corollary even two two} there are coprime \math p and \math q
such that one of them is odd and one of them is even and, 
for some \math{i\tightin\{0,1\}},
\bigmaths{p\tightsucc q},
\bigmaths{x_i\tightequal 2p q},
\bigmathnlb{x_{1-i}\tightequal p^2\tight-q^2},
and 
\ \mbox{\math{x_2\tightequal p^2\tight+q^2}}\@. \ \ 
Because of \bigmaths{x_i\tightin\posN}, we have \bigmaths{p,q\tightin\posN}. \ 
From \ \mbox{\math{x_0 x_1\tightequal 2x_3^2}}, \ 
we get \ \mbox{\math{2p q\inpit{p^2\tight-q^2}\tightequal 2x_3^2}}, \ 
\ie\ \bigmaths{p q\inpit{p^2\tight-q^2}\tightequal x_3^2}.
By \lemmref{lemma p and q coprime}, 
we know that \math{p q} and \math{p^2\tight-q^2} are coprime, too. 
Thus, by \lemmref{lemma additional one}
there must be some coprime \math{b,c\in\posN} with 
\bigmaths{x_3\tightequal b c},
\bigmaths{p q\tightequal b^2}, and
\ \mbox{\math{p^2\tight-q^2\tightequal c^2}}\@. \
By the coprimality of \math{p,q}, because of 
\ \mbox{\math{p q\tightequal b^2}}, \ 
by \lemmref{lemma additional one}
there must be some coprime \math{e,f\in\posN} with
\bigmaths{b\tightequal e f},
\bigmaths{p\tightequal e^2}, and 
\bigmaths{q\tightequal f^2}.
Moreover \bigmaths{e\tightsucc f\tightsucc 0}, as
\bigmaths{e\tightpreceq f}{} would imply the contradictory
\bigmaths{p\tightpreceq q}, and
\bigmaths{f\tightequal 0}{} would imply the contradictory
\bigmaths{x_i\tightequal 0}.
Furthermore, \ 
from 
\bigmaths{q\tightin\posN},
we get 
\bigmaths{x_2=p^2\tight+q^2\succ p^2=e^4\succeq e}.
\QED{Claim\,I}

\initial{\underline{Claim\,II:}}
There are coprime \math{g,h\in\posN} and some \math{e,f} with
\bigmaths{x_2\tightsucc e\tightsucc f\tightsucc 0}{} such that
\par\noindent\LINEnomath{\bigmaths{e^2\tight+f^2\tightequal g^2}{} and
\bigmaths{e^2\tight-f^2\tightequal h^2}.}
\initial{\underline{Proof of Claim\,II:}}
Note we will not use any information on the current proof
state beside Claim\,I here\@. \ 
By Claim\,I, \bigmaths{p\tight+q, p\tight-q\tightin\posN}.
By Claim\,I and 
\lemmref{lemma mahoney footnote one two}(2), the only prime that may divide 
both \math{p\tight+q} and \math{p\tight-q} is \nlbmath 2; \ 
but this is not the case because one of \math{p,q} is even and one is odd.
Thus, by \lemmref{lemma coprime},
\math{p\tight+q,p\tight-q} are coprime and because of \ \mbox{\math
{\inpit{p\tight+q}\inpit{p\tight-q}\tightequal p^2\tight-q^2\tightequal c^2}}, \
by \lemmref{lemma additional one}, there are coprime \math{g,h\in\posN} with
\ \mbox{\math{c\tightequal g h}}, \ 
\ \mbox{\math{p\tight+q\tightequal g^2}}, \ 
\bigmaths{p\tight-q\tightequal h^2}.
\QED{Claim\,II}

\noindent 
As 
the induction hypothesis 
in \fermat's original proof 
does not seem to be
\theoref{theorem Fermat one}, but Claim\,II instead,
let us forget anything about the current proof state 
here beside Claim\,II\@. \
The following two claims are trivial in the context of Claim\,II:

\noindent\underline{Claim\,IIa:}
\ \mbox{\math{h^2\tight+2f^2\tightequal g^2}}\@. \ 

\noindent\underline{Claim\,IIb:}
\ \mbox{\math{h^2\tight+f^2\tightequal e^2}}\@. \ 

\noindent
As \math{g,h} are coprime
(according to Claim\,II), \hskip.2em 
\math{h,f,g} are coprime, \afortiori. \
Thus, by Claim\,IIa and \cororef{corollary fermat help four},
there are \math{m,k} 
such that
\ \math{2m,k} coprime,
\ \mbox{\math{h\tightequal\CARD{2m^2\tight-k^2}}}, \ 
\ \mbox{\math{f\tightequal 2 m k}}, \ 
and 
\ \mbox{\math{g\tightequal 2m^2\tight+k^2}}\@. \ 
Set \bigmaths{y_0{:=}2m^2}, \maths{y_1{:=}k^2}, 
\bigmaths{y_2{:=}e},  
\maths{y_3{:=}m k}. \ 
By Claim\,II we have \bigmaths{x_2\tightsucc y_2\tightsucc 0}.
As \nolinebreak\bigmaths{f\tightin\posN}{} by Claim\,II, \
we have \bigmaths{m,k\tightin\posN},
and \bigmaths{y_0,y_1\tightin\posN}.
Moreover, we have
\bigmaths{y_0y_1\tightequal 2y_3^2},
\ \mbox{\math{g\tightequal y_0\tight+y_1}}, \ 
and \ \mbox{\math{f^2\tightequal 2 y_0 y_1}}\@. \ 
Finally, by Claim\,IIa and Claim\,IIb, we have 
\ \math{
y_0^2\tight+y_1^2\nottight{\nottight{=}}
\inpit{y_0\tight+y_1}^2-2y_0y_1}
{\nottight{=}
\math{
g^2-f^2\nottight{\nottight{=_{\!\!\!\!\!\!\!\!\!\mediumheadroom\rm(IIa)}}}
h^2+2f^2-f^2\nottight{\nottight{=}}
h^2+f^2\nottight{\nottight{=_{\!\!\!\!\!\!\!\!\!\mediumheadroom\rm(IIb)}}}
e^2
\nottight{\nottight{=}}
y_2^2}. \
This completes also this remaining case by {\it\descenteinfinie}.

\mbox{}\QEDbf{\theoref{theorem Fermat one}}
\vfill\pagebreak

\subsection{An Annotated Translation of \fermat's Original Proof}\label
{subsection An Annotated Translation}

\par\noindent
The following English translation of \fermat's original proof 
\hskip.1em (\cfnlb\ \sectref{section original proof}) \hskip.2em
roughly follows the translation 
found in \cite[\p\,352\f]{fermat-career}, \hskip.2em
but has several improvements. \ 
Moreover ---~to refer to the proof of \theoref{theorem Fermat one} in
\nlbsectref{subsection A Simple Self-Contained Modern Proof} explicitly~---
we have added several
annotations. \ 
(Brackets \nolinebreak \opt\ldots\ 
enclose these annotations, which are typeset in italics.) 
\halftop\begin{quote}
If the area of a \opt{\tightemph{right-angled}} triangle were a square, 
there would be given two squares-of-squares
\opt{{\math{e^4}, \math{f^4}}}
of which the difference would be a square
\opt{\tightemph{Claim\,I}}; \hskip.2em
whence it follows that two squares \opt{\maths{e^2}, 
\maths{f^2}} would be given, 
of \nolinebreak which both the sum and the difference would be squares
\opt{\tightemph{Claim\,II}}. \
And thus a number would be given,
composed of a square and the double of a square, equal to a square
\opt{\tightemph{Claim\,IIa}}, \hskip.2em
under the condition that the squares composing it make a square
\opt{\tightemph{Claim\,IIb}}.
\begin{sloppypar}
\par\halftop\noindent However, \hskip.2em 
if a square number is composed of a square and the double of another square
\opt{{\math{g^2\tightequal h^2\tight+2f^2}}}, \hskip.2em
its side \opt{{\it\ie\ its square root \nlbmath g}\hskip.05em} \hskip.2em
is similarly composed of a square and the double of a square
\opt{\math{g\tightequal k^2\tight+2 m^2}}, \hskip.2em
as we can most easily demonstrate 
\opt{\mbox{\it\cororef{corollary fermat help four}}\,}.
\end{sloppypar}
\par\halftop\noindent
Whence one concludes that this side \opt{\math g\hskip.05em} 
is the sum of the sides \opt{\math{y_0,y_1}} about the right
angle of a right-angled triangle \opt{\math{g\tightequal y_0\tight+y_1}, 
\math{y_0^2\tight+y_1^2\tightequal y_2^2}}, \hskip.2em
and that one of the squares composing \nolinebreak it 
constitutes the base
\opt{\math{k^2\tightequal y_1}}, \hskip.2em
and the double square is equal to the perpendicular
\opt{\mbox{\math{2m^2\tightequal y_0}}}.
\begin{sloppypar}
\par\halftop\noindent\opt{{\it Instead of applying \theoref{theorem Fermat one} 
as an induction
hypothesis now (or, dually, 
using its negation as the pattern for the {\it\descenteinfinie}), \hskip.2em
\fermat\ seems to descend the inductive reasoning cycle 
until the negation of Claim\,II can be applied
as an induction hypothesis
(or, dually, 
 Claim\,II can be used as the pattern for the {\it\descenteinfinie}).}} \
Hence, this right-angled triangle is generated 
\opt{{\it(in the sense of \nlbsectref{section roots one})}}
by 
two 
squares,  
of which the sum
and difference are squares. \
These \nolinebreak two \nolinebreak squares, \hskip.1em
however, \hskip.1em
will be proved to be
smaller than the first 
squares \nolinebreak initially \nolinebreak posited, \hskip.1em
of which the sum as well as
the difference also made squares: 
\par\halftop\noindent
Therefore, \hskip.1em
if two squares \opt{{\math{e^2}, \math{f^2}}} 
are given of which the sum and the difference 
are squares \opt{{\it Claim\,II}}, \hskip.2em
there exists in integers the sum of two squares of the same
nature, \hskip.1em
less \nolinebreak than the former \opt{\math{e^2\tight+f^2}}.
\end{sloppypar}
\par\halftop\noindent\opt
{\tightemph{Finally \fermat\ illustrates the Method of \DescenteInfinie.}}
By the same argument there will be given in the prior manner another
one less than this, and smaller numbers will be found indefinitely 
having the same property.
Which is impossible, because, given any integer,
one cannot give an infinite number of integers less than it.
\par\halftop\noindent
The smallness of the margin forbids to insert
the proof completely and with all detail. \
\opt{{\it This is roughly the sentence following\/ \fermatslasttheorem\
in Observation\,II \ (\cite[\Vol\,I, \p\,291]{fermat-oeuvres}, \hskip.2em 
\cfnlb\ our \sectref{section Aftermath}), \ 
which drove generations of mathematicians crazy: \
``\latintextten''}\,}\end{quote}
\vfill\pagebreak

\yestop
\subsection{Similar Interpretations of the Proof in the Literature}\label
{section Similar Interpretations of the Proof in the Literature}%
My interpretation of \fermat's proof being completed, 
it is now time to have a look into the literature of its interpretation. \ 
The interpretation of \fermat's proof in \cite[\Vol\,2, \p\,615\f]{dickson}, \ 
(looked up only after my interpretation was already completed) \ 
is \nolinebreak roughly similar to my interpretation, 
but a little less structured and less similar to \fermat's original proof. \ 
The interpretation of the proof in 
\cite[\litsectref{1.6}]{fermat-last-theorem}, \ 
(looked up only after my interpretation was already completed) \ 
claims to follow \hskip.1em\cite[\Vol\,2, \p\,615\f]{dickson}, \ 
but makes things only worse and is not at all convincing.
\begin{quote}
``It is the next two sentences 
\opt{\tightemph{our \nth 2 and \nth 3 paragraphs}\/} 
that are the difficult ones to follow 
\opt{\tightemph{for Edwards}}.''\getittotheright
{\cite[\litsectref{1.6}, \p\,13]{fermat-last-theorem}}\end{quote}
\par\noindent
The briefest interpretation of the proof in 
\cite[\litchapref{X}]{weil-number-theory} \hskip.2em
(looked up only after my interpretation was already completed) \hskip.2em
is quite in accordance with my interpretation
when we apply the substitution
\maths{\{
x\tight\mapsto e,
y\tight\mapsto f,
u\tight\mapsto g,
v\tight\mapsto h,
z\tight\mapsto c,
r\tight\mapsto k,
s\tight\mapsto m\}}{} to \weil's interpretation.
\par\yestop\noindent
The interpretation of the proof in 
\cite[\litchapref{VI.VI}]{fermat-career} \hskip.2em 
is less brief, but mathematically strange in the sense that there are
some steps in it which I do not clearly understand \ 
(\mbox{such as} ``we may set \math{f^2\tightequal 4 k^2m^2}.'')\@. \ 
As \cite{fermat-career} is the standard work on the mathematics of \fermat,
however, 
we have renamed our variables in accordance to it, 
such that the  
proof of \cite{fermat-career} is already 
roughly in accordance with our \mbox{presentation here}, \hskip.1em
without any renaming of variables.
\par\yestop\noindent
The discussion of the proof in 
\cite[\litchapref{2.2.3}, \PP{39}{46}]{From-Fermat-to-Gauss}
follows the interpretation of \cite{fermat-career}, \ 
but elaborates the 
ways and means of the induction-hypothesis application, or more precisely,
the indefinite descent. \ 
We read:\begin{quote}``From this demonstration, it is possible to deduce one
of the most important properties inherent in every argument by indefinite
descent: there is an\emph{invariable form} with different orders of sizes 
\opt{\ldots}.
\getittotheright{\cite[\p\,45]{From-Fermat-to-Gauss}}
\end{quote}
As we have already discussed in \sectref{subsection A Further Refinement}, \ 
from a refined logic-historical point of view, this 
``\tightemph{invariable form}'' may actually vary. \ This will become important
in \sectref{section Interpretation} below, where we will discuss  
the only interpretation of \fermat's original proof that significantly differs
from my presentation here, namely the one in \cite{walsh-fermat-XLV}.
\vfill\pagebreak
\yestop
\subsection{\walsh's Alternative Interpretation}\label{section Interpretation}
The only interpretation of \fermat's original proof that significantly differs
from our presentation here is the one in \cite{walsh-fermat-XLV}. \ 
With the missing details added and the unconvincing parts removed, 
we may describe \walsh's interpretation roughly as follows.
\par\yestop\noindent
Suppose that the clean-up of 
\sectref{subsection A Simple Self-Contained Modern Proof}
up to Claim\,I has been done.
Then we continue as follows.
\par\halftop\noindent\underline{Claim\,III:} \ 
There are 
some \math{e,f,g,h\in\posN} 
with \math{g,h} coprime and
\bigmaths{x_2\tightsucceq e\tightsucc f\tightsucc 0}{} such that
\par\noindent\LINEnomath{\bigmaths{e^2\tight+f^2\tightequal g^2}{} and
\bigmaths{e^2\tight-f^2\tightequal h^2}.}
\par\noindent\underline{Proof of Claim\,III:} \ \ 
Set 
\bigmaths{e:=x_2}, 
\bigmaths{f:=2x_3},
\bigmaths{g:={x_0\tight+x_1}}, and
\bigmaths{h:=\CARD{x_0\tight-x_1}}.
Then we have 
\bigmaths{e^2\tight\pm f^2
\tightequal x_2^2\tight\pm 4x_3^2
\tightequal x_0^2\tight+x_1^2\tight\pm 2x_0x_1
\tightequal\inpit{x_0\tight\pm x_1}^2}.
Moreover, we have \bigmaths{e\tightsucc f}{} by the following indirect proof:
Otherwise, we would have
\bigmaths{0\tightsucceq e^2\tight-f^2\tightequal\inpit{x_0\tight-x_1}^2},
\ie\ \bigmaths{x_0\tightequal x_1}, and 
as \math{x_0, x_1} are coprime, we would get \ \mbox{\math
{x_0\tightequal x_1\tightequal 1}}, \ \ie\ the contradictory
\bigmaths{2\tightequal x_2^2}. \ 
Finally, to show that \math{g,h} are coprime suppose
\bigmaths{z\mid x_0\tight+x_1}{} and \bigmaths{z\mid\CARD{x_0\tight-x_1}}.
As \math{x_0,x_1} are coprime, we get \bigmaths{z\tightpreceq 2}{}
by \lemmref{lemma mahoney footnote one two}(2). \ 
By \cororef{corollary even two two} we know that \math{x_0\tight+x_1}
is odd. Thus, we get \bigmaths{z\tightequal 1}, as was to be shown.
\QED{Claim\,III}

\yestop\noindent
Compared to Claim\,II of 
\sectref{subsection A Simple Self-Contained Modern Proof},
the weakness of 
Claim\,III is that it only states 
\bigmaths{x_2\tightsucceq e}{}
instead of \bigmaths{x_2\tightsucc e}.
This weaker statement, however, does not admit us to apply our
induction hypothesis as before. 
This is not by chance and another 
weight function cannot help us,
because the new triangle is actually the same as before:
Indeed, we have
\ \mbox{\math{x_2\tightequal e\tightequal y_2}} \ and
\bigmaths{\{x_0,x_1\}=
\{{g\tight\pm h\over 2}\}=
\{2m^2,k^2\}=
\{y_0,y_1\}
}. \ 
This means that
---~to arrive in proof state with a smaller weight~---
we actually have to descend the inductive reasoning
cycle by proving Claim\,I and Claim\,II\@. \ 
The interesting aspect is that
---~as noted already in \sectref{subsection An Annotated Translation}~---
this is exactly what \fermat\ does in his proof.
While these steps are superfluous
according to all other interpretations, 
they are necessary according to the interpretation in \cite{walsh-fermat-XLV}.
\par\yestop\noindent
Note that, \hskip.1em
according to \sectref{subsection A Further Refinement}, \hskip.2em
\walsh's proof is actually sound \wrt\ the following instantiation: \ 
In \nlbmath{\inpit{\ident{ID}'}} 
of \nolinebreak\sectref{subsection A Further Refinement}, \
roughly speaking, \ 
we set \math{P_0} to \theoref{theorem Fermat one}, \ and 
\math{P_i} to Claim\,II for all \math{i\in\posN}. \ 
We measure \math{P_0} with the weight 
\nlbmath{x_2^2\tight+\inpit{2x_3}^2\tight+1}. \ 
And we measure \math{P_i} with the weight \nlbmath{e^2\tight+f^2} 
for \math{i\in\posN}, \ 
just as \fermat\ has described the weight in his original proof,
\cfnlb\ the \nth 4 paragraph of our annotated translation in 
\nolinebreak\sectref{subsection An Annotated Translation}. \ 
From \math{P_0} to \nlbmath{P_1} the weight decreases by 1\@. \ 
And from \math{P_i} to \math{P_{i+1}} the weight also
decreases for \math{i\in\posN}. \ \ 
A \nolinebreak simpler way to see the soundness of this proof 
is to model it as a deductive 
proof of \theoref{theorem Fermat one} with Claim\,II as a lemma,
plus an inductive proof of Claim\,II\@.
\par\yestop\noindent
We cannot decide whether the interpretation of 
\citet{walsh-fermat-XLV} \hskip.1em
for the first step of \fermat's proof 
\hskip.1em (\ie\ the Proof of Claim\,III) \hskip.2em
reflects \fermat's intentions regarding his original proof 
better than our presentation in \sectrefs
{subsection A Simple Self-Contained Modern Proof}
{subsection An Annotated Translation}. \hskip.3em 
The interpretation of the actual indefinite descent of \fermat, \hskip.2em 
however, \hskip.2em
is superior in \walsh's version. \hskip.2em
Moreover, \hskip.2em
note that
\mbox{---~considering \fermat's extreme conciseness in general~---}
it is very likely that \fermat\ had believed the additional
descent to be actually necessary. \ 
We cannot decide, however, whether this belief, 
which he contradicts in his letter for \huygens\ 
in\,\letterforhuygensyear\ 
\hskip.1em (\cfnlb\ \sectref{section French statement}), \ 
means that \walsh's interpretation is the correct one. \ 
This belief may be simply based on \fermat's
inexperience with\emph\descenteinfinie\ at the time when he wrote his
\OBS, more than 20 years before his death in\,\fermatdeathyear. \ \ 
The following dating of the Observations very roughly agrees with 
\cite[\litsectref{11}]{catherine-goldstein}:
\begin{quote}``The dates of the
various notes on \diophantus\ are not known; but it is probable that this
Note \opt{\tightemph{\OBS}}
was written sometime between 1636 and 1641, or at least, in round 
numbers twenty years before the Letter \opt{\tightemph{for \huygens}}. \ 
This is important; for the Note and the Letter do not agree
---~at least in appearance 
\opt{\tightemph{of the induction-hypothesis application%
}}.'' \getittotheright{\cite[\p\,412]{walsh-fermat-XLV}}\end{quote}
\yestop\yestop
\subsection{\frenicle's More Elegant Version of \fermat's Proof}\label
{section aside}
Note that geometric illustration cannot help much to understand 
the Proof of \theoref{theorem Fermat one} in 
\sectref{subsection A Simple Self-Contained Modern Proof}. \ 
The following seems to be the best we can get:\par\noindent
\LINEmath{\xymatrix{
&&&&{}\ar@{-}[dddd]^{\begin{array}{@{}l@{}}
  \scriptstyle x_{1-i}\tightequal
\\\scriptstyle c^2\tightequal
\\\scriptstyle\inpit{p\tight+q}\inpit{p\tight-q}\tightequal
\\\scriptstyle g^2 h^2\tightequal
\\\scriptstyle\inpit{2m^2\tight+k^2}^2 \inpit{2m^2\tight-k^2}^2
\\\end{array}}
\ar@{-}[lllldddd]_{x_2\tightequal y_2^4+\inpit{2 m k}^4\tightequal p^2+q^2}
&&&&&{}\ar@{-}[dr]^{c\tightequal g h
\tightequal\inpit{2m^2\tight+k^2}\CARD{2m^2\tight-k^2}}
\ar@{-}[lddd]_{p\tightequal e^2}
\\&&&&&&&&&&{}\ar@{-}[ddll]^{q\tightequal\inpit{2m k}^2}
&&&
\\&&&&&{}&&&&&&&&{}\ar@{-}[dd]^{y_0\tightequal 2m^2}
\ar@{-}[ldd]_{y_2\tightequal e}
\\&&&&{}&{}&{}&{}&{}&{}&{}&{}&{}
\\{}\ar@{-}[rr]_{p q}
&&{}\ar@{-}[rr]_{p q\tightequal y_2^2\inpit{2 m k}^2}
&&&&&&&&{}&&{}\ar@{-}[r]_{y_1\tightequal k^2}&{}
}}\par\noindent
The leftmost triangle is the originally assumed one and the rightmost
triangle is the one to which the modern Proof of \theoref{theorem Fermat one} 
in \sectref{subsection A Simple Self-Contained Modern Proof} descends. \ 
Although the one in the middle is rectangular by Claim\,I, \ 
it is not explicitly noted in \fermat's proof. \ 
In the proof of the same theorem in \cite{frenicle}, however,
\freniclename\ \mbox\freniclelifetime\
(\Paris)
descends to the same rightmost triangle
but completely avoids application of \cororef{corollary fermat help four}
by an application of \lemmref{lemma frenicle XXXVIII}
to the rectangular triangle depicted in the middle. \ 
Omitting Claim\,II and all the following, \ 
the proof of \litpropref{XXXIX} in \cite{frenicle}
continues roughly as follows: \ 
We have (Claim\,I)
\bigmaths{q^2\tight+c^2\tightequal p^2}, \math c is odd,
\bigmaths{p\tightequal e^2}, 
\ \mbox{\math{q\tightequal f^2}}, 
\bigmaths{x_2\tightsucc e\tightsucc f\tightsucc 0},
and \math{p,q} are coprime.
By \lemmref{lemma frenicle XXXVIII},
there are \math{m,k\in\posN}
such that 
\ \mbox{\math{\headroom\footroom
\inpit{2m^2}^2\tight+\inpit{k^2}^2\tightequal p\tightequal e^2}. \
\qedabbrev}\ \ \
This is more elegant than \fermat's proof with an obviously
and definitely different lemmatization, which, however, 
was sometimes neglected:
\begin{quote}
``\frenicle\ follows this proof \opt{\tightemph{of \fermat}\,} faithfully,
with little more than verbal changes \opt{\ldots}.\getittotheright
{\cite[\litsectref X, \p\,77]{weil-number-theory}}\end{quote}
As we are already exceeding the scope of this little \daspaper\ 
and our margins are too small, we ask the reader who is interested in more
information on the subject to have a look at
\cite{catherine-goldstein}, \
which is a whole book dedicated to the
history of \theoref{theorem Fermat one}, \
including a discussion of different interpretations
of \fermat's Latin original and much more.
\vfill\cleardoublepage
\yestop\section{Conclusion}
\par\noindent
\fermat\ gave his readers a hard time with his notes. \
His proof sketch of his \OBS, \hskip.2em 
which we have studied in this \daspaper\ in some detail, \hskip.1em
is hard to understand, interpret, and disambiguate; \hskip.3em
for \nolinebreak the readers of the \nth{21}\,century just as well
as for readers of \nth{17}\,century. \
Without expertise in number theory, 
it takes some days to construct a consistent interpretation of
this short proof sketch.
\par\halftop\indent
\fermat\ named the method of this proof {\it\descenteinfinie}. \
This method and its variants are of outstanding importance in mathematics. \
\par\halftop\indent
In \sectref{section descente infinie}, \hskip.2em
we have discussed the Method of {\it\DescenteInfinie}\/
from the mathematical,
\mbox{logical}, historical, linguistic, and refined logic-historical 
\mbox{points of view}, \hskip.2em
and we have presented all its aspects with novel 
clearness, precision, and detail.
\par\halftop\indent
As \fermat\ wanted people to have fun with number theory, \hskip.2em
we \nolinebreak have suggested that our
\sectref{section Prerequisites from Number Theory}
could be skipped by the readers and have put its less interesting
proofs into the appendix (\sectref{section missing proofs}); \hskip.3em
so a reader who is not an expert in number theory may choose between 
the fun of exercise or the relief of solution.
\par\halftop\indent
Regarding the aspects of both pedagogical presentation 
and the interpretation of critical texts in the 
history of mathematics, \hskip.1em
we consider it to be advantageous to 
present the following three in the given order:\begin{enumerate}\noitem\item 
the Latin original proof \nolinebreak 
(\sectref{section original proof}), \noitem\item 
a modern self-contained proof \nolinebreak
(\sectref{subsection A Simple Self-Contained Modern Proof}), \noitem\item 
and an English translation of the Latin original \nolinebreak
(\sectref{subsection An Annotated Translation}).\noitem\end{enumerate}%
This \daspaper\ is
(to the best of our knowledge) \hskip.1em
unique already in presenting
these three items. \ 
\par\halftop\halftop\indent
Moreover, \hskip.2em
this \daspaper\ is (again to the best of our knowledge) \hskip.1em
unique in annotating 
the English translation with references to a more explicit modern proof
and not vice versa. \
Although we have been quite laconic with our comments in the
translation, \hskip.2em
we are confident that 
the mathematical gestalt of \fermat's proof sketch is perceivable 
with the help of the sparse
annotations in the translation 
in \nolinebreak\sectref{subsection An Annotated Translation}, \hskip.2em
building on the detailed presentation of the modern proof
in \nolinebreak\sectref{subsection A Simple Self-Contained Modern Proof}. \ 
We believe that this perception is easier and deeper 
than what can arise from the standard procedure 
of presenting a modern proof with annotations from
(a \nolinebreak translation \nolinebreak of) the 
original, \hskip.2em
and that the usefulness of our procedure 
for interpreting critical texts in the 
history of mathematics is higher than that of the 
standard procedure.
\par\halftop\indent
All in all, \hskip.2em
including all important facts, \hskip.2em
we have presented a concise and self-contained 
discussion of \fermat's proof sketch, \hskip.2em
which is easily accessible to laymen in number theory as well as
to laymen in the history of mathematics, \hskip.2em
and which provides new clarification of the 
Method of {\it\DescenteInfinie}\/ to the experts in these fields. \
Last but not least, \hskip.2em
this \daspaper\ fills a gap regarding the 
easy accessibility of the subject.
\vfill\pagebreak

\yestop\yestop\section{Aftermath}\label{section Aftermath}
Finally, note that \fermat's proof sketch omits the little but important 
details of the proof,
such as being positive, being coprime, or \bigmaths{2\nmid p\tight-q},
which are essential parts of our modern proof in 
\sectref{subsection A Simple Self-Contained Modern Proof}. \ 
Moreover, he does not at all explicate the underlying theory,
which we tried to reconstruct in 
\sectref{section Prerequisites from Number Theory}. \ 
\par\halftop\indent
As he seems to have found his proofs without
pen and paper just in his imagination, he may have had some subconscious
subroutines taking care of this.
Such subroutines are error-prone and would explain \fermat's 
only claim 
that we know to be wrong, namely\newcommand
\thenonprimeprime[1]{2^{2^#1}\tight+1}
to have proved \bigmaths{\forall n\tightin\N\stopq\inpit
{\thenonprimeprime n\mbox{ prime}}}, contradicted by
\ \mbox{\math{5=\mu n\stopq\neg\inpit{\thenonprimeprime n\mbox{ prime}}}} \ 
because of
\bigmaths{2^7 5\tight+1\mid\thenonprimeprime 5}. \ 
This claim of a proof occurs in that same letter for \huygens, 
which we quoted in \sectref{section French statement} and discussed
in \sectref{section Revival}. \
Five \nolinebreak years before, in a letter to \pascal\ in August\,1654,
he had admitted that 
the proof was still incomplete; \hskip.3em
\cfnlb\ \cite[\Vol\,II, \p\,309\f]{fermat-oeuvres}. \ 
\par\halftop\indent
Whether the proof \fermat\ 
claimed to have found for ``\fermat's Last Theorem''
\par\noindent\LINEmaths{
\forall n\tightsucceq 3\stopq
\forall x,y,z\tightin\posN\stopq\inpit{x^n\tight+y^n\tightnotequal z^n}}
{}\par\noindent
was also faulty for bigger \nlbmath n,
or whether \fermat, the methodologist
who after eighteen centuries was the first to apply 
the Method of\emph\Descenteinfinie\ again, 
invented yet another method for this proof, 
is still an open question.

\yestop\yestop
\section*{Acknowledgments}\addcontentsline{toc}{section}{Acknowledgments}
I would like to thank \bussottiname\ for a 
list of suggestions for improvement on an earlier version of this \daspaper,
for some profound help, 
and for much more.

I would also like to thank \goldsteinname\ for some information 
on the sophistication of the 
publications of L'Acad\'emie Royale des Sciences, Paris. \

Furthermore, I would like to thank \stephanunverzagtname\
for some alternative translations of \fermat's Latin text,
which were especially useful because they did not take the mathematical content
into account.

Finally, I would like to thank \wagnername\ for some help with the
French language, and for much more.

\vfill\pagebreak

\appendix
\yestop
\section{Missing Proofs and Additional Lemmas}\label
{section missing proofs}%

\yestop\begin{proofparsepqed}{\lemmref{lemma Euclid VII 20}}
Let us assume \bigmaths{x_0 y_1\tightequal y_0 x_1}{} with 
\math{x_0,x_1,y_0,y_1\in\posN}.
\initial{\underline{(i):}}
We show the ``only if''-direction, the ``if''-direction is symmetric.
\\\bigmaths{
x_0 z_1 y_0
\tightequal
{x_0 y_0 z_1}
\tightequal
{x_0 z_0 y_1}
\tightequal
{z_0 x_0 y_1}
\tightequal 
{z_0 y_0 x_1}
\tightequal 
z_0 x_1 y_0
}.
Thus, dividing by \nlbmath{y_0},
we get \bigmaths{x_0 z_1\tightequal z_0 x_1}.

\initial{\underline{(iii)\tightimplies(ii):}}
Assume (iii). \
If \math{y_0,y_1} were not coprime, there would be some
\math{k\tightsucceq 2} and \math{y_i''\in\posN} with 
\bigmaths{k y_i''\tightequal y_i}{} for \math{i\in\{0,1\}}. \
Then we would have \bigmaths{x_0 k y_1''\tightequal k y_0'' x_1},
and then \bigmaths{x_0 y_1''\tightequal y_0'' x_1}.
This would contradict the \tightpreceq-minimality of \nlbmath{y_0}.

\initial{\underline{(iii)\tightimplies(iv):}}
Assume (iii). \
Dividing \math{x_i} by \math{y_i}, \hskip.2em
we get \math{k_i,r_i\in\N} with
\bigmaths{x_i\tightequal k_i y_i\tight+r_i}{} and
\bigmaths{r_i\tightprec y_i}{} for \math{i\in\{0,1\}}. \ 
Then \bigmaths{k_0 y_0 y_1 + r_0 y_1
=\inpit{k_0 y_0 + r_0} y_1
=x_0 y_1
=y_0 x_1
=y_0\inpit{k_1 y_1 + r_1}
=k_1 y_0 y_1 + r_1 y_0}{}
with \bigmaths{r_i y_{1-i}\tightprec y_0 y_1}{} for \math{i\in\{0,1\}}. \ 
Thus, \hskip.1em
dividing \math{x_0 y_1} by \math{y_0 y_1}, \hskip.2em
we get
\math{k_0} with remainder \nlbmath{r_0 y_1} as well as
\math{k_1} with remainder \nlbmath{r_1 y_0}. \
Thus, \ \mbox{\math{k_0\tightequal k_1}} \ 
and \bigmaths{r_0 y_1\tightequal r_1 y_0}.
Then \bigmaths{x_0 r_1 y_0=x_0 r_0 y_1=x_0 y_1 r_0=y_0 x_1 r_0=r_0 x_1 y_0}.
Thus, we have \bigmathnlb{x_0 r_1\tightequal r_0 x_1}.
But as \bigmaths{r_0\tightprec y_0}{} and as
\math{y_0} is the least number 
such that there is a \math{y_1'\in\posN} with
\bigmathnlb{x_0 y_1'\tightequal y_0 x_1},
we have \bigmathnlb{r_0\tightequal 0}. 
This implies \bigmathnlb{r_1\tightequal 0}.
Thus, \bigmaths{x_i\tightequal k_0 y_i}{}  
for \math{i\in\{0,1\}}.

\initial{\underline{(ii)\tightimplies(iii):}}
Assume (ii). \
As \bigmaths{y_0 z_1\tightequal z_0 y_1}{} 
has the solution \bigmaths
{\pair{z_0}{z_1}\tightequal\pair{x_0}{x_1}\tightin\posN\tighttimes\posN}, 
let \math{z_0} be \tightpreceq-minimal 
such that there is a \math{y_1'\in\posN} with
\bigmathnlb{y_0 y_1'\tightequal z_0 y_1}.
Then \nolinebreak\bigmaths{z_0\tightin\posN}.
Let \math{z_1\in\posN} be given such that 
\bigmathnlb{y_0 z_1\tightequal z_0 y_1}.
Applying (iii)\tightimplies(iv) to the sentence 
\bigmathnlb{y_0 z_1\tightequal z_0 y_1},
we infer that 
there is a \math{k\in\posN} such that \bigmathnlb{k z_i\tightequal y_i}{}
for \math{i\in\{0,1\}}. \
Thus, \bigmaths{k\mid y_i}{} for \math{i\in\{0,1\}}. \
As \math{y_0,y_1} are coprime, we have
\ \mbox{\math{k\tightequal 1}}. \ 
Thus, \bigmaths{z_0\tightequal y_0}.
Thus, \math{y_0} is \tightpreceq-minimal such that there is 
a \math{y_1'} with \bigmathnlb{y_0 y_1'\tightequal y_0 y_1}.
By \nolinebreak (i),
\math{y_0} is \tightpreceq-minimal such that there is 
a \math{y_1'} with \bigmathnlb{x_0 y_1'\tightequal y_0 x_1}.
\end{proofparsepqed}

\yestop\begin{proofqed}{\lemmref{lemma Euclid VII 23}} \ \ 
Assume \bigmaths{u\mid x}{} and \bigmaths{u\mid z}.
By transitivity of \nlbmath\mid\ according to \cororef{corollary divides},
from \bigmaths{x\mid y}{} we have \bigmaths{u\mid y}. 
From \math{y,z} being coprime we get \bigmaths{u\tightequal 1}.\end{proofqed}

\yestop\begin{proofqed}{\lemmref{lemma Euclid VII 24}} \ \ 
For \math{m\tightequal 0}, the lemma holds as \math 1 is the minimal
element of the reflexive ordering \nlbmath\mid\ according to 
\cororef{corollary divides}. 
Thus, let us show that the lemma holds for \nlbmath{m\tight+1}
under the induction hypothesis that it holds for \nlbmath m. \
If there is some \math{i\in\{1,\ldots,m\tight+1\}} with 
\ \mbox{\math{x_i\tightequal 0}}, \ 
then we have \bigmaths{z\tightequal 1}, and the lemma holds 
for \nlbmath{m\tight+1}. \ 
In case of \bigmathnlb{z\tightequal 0}, we have \bigmaths{x_i\tightequal 1}{}
for all \math{i\in\{1,\ldots,m\tight+1\}}, \hskip.2em
and the lemma holds again for \nlbmath{m\tight+1}. \ 
Thus we may assume \bigmaths{x_0,\ldots,x_{m+1},z\tightin\posN}.
Assume \bigmaths{u\mid\prod_{i=1}^{m+1} x_i}{} and \bigmaths{u\mid z}.
By the first there is some \nlbmath{k'} with 
\bigmaths{x_{m+1}\prod_{i=1}^{m} x_i\tightequal k' u}{}
and \bigmaths{k',u\tightin\posN}.
By the second, by \math{x_{m+1},z} being coprime, 
and by \lemmref{lemma Euclid VII 23},
we get that \math{x_{m+1},u} are coprime.
Thus, by \lemmref{lemma Euclid VII 21}, there is some \math{k\in\posN} with
\bigmaths{k u\tightequal\prod_{i=1}^m x_i}.
But then \bigmaths{u\mid\prod_{i=1}^m x_i}. 
By our induction hypothesis, \bigmaths{\prod_{i=1}^m x_i\comma z}{} 
are coprime. \
Thus, we get \bigmaths
{u\tightequal 1}.\end{proofqed}
\pagebreak

\yestop\begin{proofqed}{\lemmref{lemma Euclid VII 26}} \ \ 
For \math{m\tight+n\tightequal 0}, the lemma holds as \math 1 is the minimal
element of the reflexive ordering \nlbmath\mid\ according to 
\cororef{corollary divides}. Thus, suppose the lemma holds for 
arbitrary \nlbmath{m\tight+n} to show that it holds for 
\nlbmath{m\tight+n\tight+1}.
By symmetry, we may assume that \math{x_i,y_j} are coprime for\/ 
\math{i\in\{1,\ldots,m\}}{}
and\/ \math{j\in\{1,\ldots,n\tight+1\}}. \ 
By \lemmref{lemma Euclid VII 24} 
\bigmaths{y_{n+1}\comma\prod_{i=1}^m x_i}{} are coprime.
By induction hypothesis, 
\bigmaths{\prod_{i=1}^n y_i\comma\prod_{i=1}^m x_i}{} are coprime.
All in all, by \lemmref{lemma Euclid VII 24} 
\bigmaths{y_{n+1}\prod_{i=1}^n y_i\comma\prod_{i=1}^m x_i}{} are coprime,
\ie\ 
\bigmaths{\prod_{i=1}^m x_i\comma\prod_{i=1}^{n+1} y_i}{} are coprime.
\end{proofqed}

\yestop\begin{proofqed}{\lemmref{lemma Euclid VII 29}} \ \ 
Assume \bigmaths{k\mid p}{} and \bigmaths{k\mid x}.
As \math p is prime, we have \bigmaths{k\tightin\{1,p\}}.
As \bigmaths{p\nmid x}, we have \bigmaths{k\tightnotequal p}.
Thus, \bigmaths{k\tightequal 1}.
\end{proofqed}

\yestop\begin{proofqed}{\lemmref{lemma Euclid VII 30}} \ \ 
We may assume \bigmaths{p\nmid x_1}{} and \bigmaths{x_0,x_1\tightin\posN}.
Then \math{p,x_1} are coprime by \lemmref{lemma Euclid VII 29}. \ 
Because of \bigmaths{p\mid x_1 x_2}, there is some \math k with 
\bigmaths{k p\tightequal x_1 x_2}.
Then \bigmaths{k\tightin\posN}.
By \lemmref{lemma Euclid VII 21}, we get \bigmaths{p\mid x_2}.\end{proofqed}

\yestop\begin{proofparsepqed}{\lemmref{lemma Euclid VII 31}}
We show this by {\it\descenteinfinie},
more precisely by indefinite descent (ID, \cfnlb\ \p\pageref{page ID}).
\par
Suppose that there is some \math{x\tightnotequal 1} not divided by any prime.
We look for an \math{x'\tightnotequal 1} not divided by any prime
with \bigmaths{x'\tightprec x}.
As the relation \nlbmath\mid\ is reflexive and has maximum \nlbmath 0
according to \cororef{corollary divides}, \ 
\math x is not prime and \bigmaths{x\tightsucceq 2}. 
Thus, there must be
some \math{x'\notin\{1,x\}} with \bigmaths{x'\mid x}.
As the relation \nlbmath\mid\ is transitive 
according to \cororef{corollary divides}, \ 
\math{x'} \nolinebreak is not divided by any prime.
Moreover, there is a \math k such that \bigmaths{k x'\tightequal x}.
From \bigmaths{x\tightsucceq 2}{} and \bigmaths{x'\tightnotin\{1,x\}},
we thus get \bigmaths{k\tightsucceq 2}{} and\bigmaths{x'\tightsucceq 2}.
Thus, \bigmaths{x'\tightprec x}.\end{proofparsepqed}

\yestop\begin{proofparsepqed}{\lemmref{lemma Euclid VIII 1}} 
\underline{Claim\,1:} \ \bigmaths{x_0 y_i\tightequal y_0 x_i}{}
and \bigmaths{x_{i-1} y_i\tightequal y_{i-1} x_i}{}
for all \math{i\in\{1,\ldots,n\tight+1\}}. 
\\\underline{Proof of Claim\,1:} \ 
For \math{i\tightequal 1} this holds by assumption of the
lemma. Suppose it holds for \math{i\in\{1,\ldots,n\}}. \
Then \bigmaths{x_0 y_{i+1}
\tightequal\frac{x_0 y_i^2}{y_{i-1}}
\tightequal\frac{y_0 x_i y_i}{y_{i-1}}
\tightequal\frac{y_0 x_i^2}{x_{i-1}}
\tightequal y_0 x_{i+1}
}{}
and \ \mbox{\math{x_{i} y_{i+1}
\tightequal\frac{x_{i} y_i^2}{y_{i-1}}
\tightequal\frac{x_{i}^2 y_i}{x_{i-1}}
\tightequal y_i x_{i+1}}}.\QED{Claim\,1}
\par\noindent 
From Claim\,1 we get \bigmaths{x_0 y_{n+1}\tightequal y_0 x_{n+1}}.
As \math{x_0,x_{n+1}} are coprime,
by \lemmref{lemma Euclid VII 21} there is some \math{k\in\posN}
with \bigmaths{k x_i\tightequal y_i}{}
for \math{i\in\{0,n\tight+1\}}. \
By Claim\,1 this holds for all \math{i\in\{0,\ldots,n\tight+1\}}
because of \ \mbox{\math{k x_i\tightequal\frac{k x_0 y_i}{y_0}\tightequal
\frac{y_0 y_i}{y_0}\tightequal y_i}}.\end{proofparsepqed}

\yestop\begin{proofqed}{\lemmref{lemma Euclid VIII 2}} \ \ 
Let \math{y} be minimal with
\ \mbox{\math{x_0 z\tightequal y x_1}} \ 
for some \math{z\in\posN}. \ 
By 
\lemmref{lemma Euclid VII 20}, 
\bigmaths{y\tightin\posN}, \ 
\math{y,z} are coprime, and 
there is some \math{k'\in\posN} with
\bigmaths{k' y\tightequal x_0}{} and \ \mbox{\math{k' z\tightequal x_1}}. \ 
Thus, 
\bigmaths
{x_0\inpit{y^n z^1}\tightequal k' y^{n+1} z^1\tightequal\inpit{y^{n+1} z^0}x_1}.
Moreover, \math{y^{n+1} z^0, \ldots, y^{n+1-i} z^{i},\ldots,y^0 z^{n+1}}
are in continued proportion. By \cororef{corollary Euclid VII 27},
\math{y^{n+1},z^{n+1}} are coprime, too. 
Applying \lemmref{lemma Euclid VIII 1}, we get a \math{k\in\posN} with
\bigmaths{k y^{n+1-i} z^{i}\tightequal x_i}{}
for\/ \math{i\in\{0,\ldots,n\tight+1\}}. \ 
In case that \math{x_0,x_{n+1}} are coprime, we get by
\lemmref{lemma Euclid VIII 1} a \math{k''\in\posN} with
\ \mbox{\math{k'' x_i\tightequal y^{n+1-i} z^{i}}}. \
This means \bigmaths
{x_i\tightpreceq y^{n+1-i} z^{i}},
\ie\ \bigmaths{k\tightequal 1}.\end{proofqed}
\pagebreak

\noindent
Below we will prove \litpropref{VIII.14} of 
\euclid's Elements, but we give a proof that is shorter and more modern 
than the one in the Elements, which 
recursively requires a large number of additional propositions.
Instead we need the following two lemmas.

\yestop\noindent The following lemma will be applied exclusively in
the proofs of \lemmrefs{lemma divides}{lemma Euclid VIII 14}.
\begin{lemma}\label{lemma new one}
If\/ \math p is prime and \bigmaths{p^m\mid x_0 x_1}, then there are 
\math{n_0} and \math{n_1} such that 
\ \mbox{\math{m\tightequal n_0\tight+n_1}}, \ 
\bigmaths{p^{n_0}\mid x_0}, and \bigmaths{p^{n_1}\mid x_1}.\end{lemma}
\begin{proofparsepqed}{\lemmref{lemma new one}}
We show this by {\it\descenteinfinie},
more precisely by indefinite descent (ID, \cfnlb\ \p\pageref{page ID}).
\par
Let \math p be prime.
Suppose that \bigmaths{p^m\mid x_1 x_2}, but there are no 
\math{n_1} and \math{n_2} such that 
\ \mbox{\math{m\tightequal n_1\tight+n_2}}, \ 
\bigmaths{p^{n_1}\mid x_1}, and \bigmaths{p^{n_2}\mid x_2}.\
Then \bigmaths{m,x_0,x_1\tightin\posN}.
By \lemmref{lemma Euclid VII 30}, there is an \math{i\in\{0,1\}}
such that \bigmaths{p\mid x_i}.
Thus, there is some \math{x'_i\in\posN} with \bigmaths{x'_i p\tightequal x_i}.
Set \math{x'_{1-i}:=x_{1-i}}. \ 
There is some \math{k\in\posN} with \bigmaths{k p^{m}\tightequal x_0 x_1}.
Thus, \bigmaths{k p^{m}\tightequal x'_{1-i}x'_i p}.
Thus, \bigmaths{k p^{m-1}\tightequal x'_{1-i}x'_i}. 
But there cannot be any \math{n'_1} and \math{n'_2} such that 
\ \mbox{\math{m\tight-1\tightequal n'_1\tight+n'_2}}, \ 
\bigmaths{p^{n'_1}\mid x'_1}, and \bigmaths{p^{n'_2}\mid x'_2},
because this leads to a contradiction when we set
\math{n_i:=n'_i\tight+1} and \math{n_{1-i}:=n'_{1-i}}.\end
{proofparsepqed}

\yestop\yestop\noindent 
The following lemma will be applied exclusively in the proof
of \lemmref{lemma Euclid VIII 14}.
\begin{lemma}\label{lemma divides} \ 
\bigmaths{x\mid y}{} \uiff\ \bigmaths{\forall p\mbox{ prime}\stopq
\forall n\tightin\posN\stopq\inparentheses
{\inpit{p^n\mid x}\nottight{\nottight\implies}\inpit{p^n\mid y}}}.\end{lemma}
\begin{proofqed}{\lemmref{lemma divides}} \ \ 
The ``only if''-direction follows directly from the transitivity 
according to \cororef{corollary divides}.
We show the other direction by\emph\descenteinfinie, 
more precisely by indefinite descent:
Suppose that there are \math{x,y} such that
\bigmaths{\forall p\mbox{ prime}\stopq
\forall n\tightin\posN\stopq\inparentheses
{\inpit{p^n\mid x}\nottight{\nottight\implies}\inpit{p^n\mid y}}},
but \bigmaths{x\nmid y}.
We find \math{x',y'} of the same kind with \bigmaths{y'\tightprec y}.
According to \cororef{corollary divides}, we have
\bigmaths{x\tightnotequal 1}{} and \bigmaths{y\tightnotequal 0}.
By \lemmref{lemma Euclid VII 31}, 
there is some prime \nlbmath p such that \bigmaths{p\mid x}.
Then we have \bigmaths{p\mid y}{} by our assumption (setting \math{n:=1}). \
Thus, we have \bigmaths{y\tightsucceq 2}.
Moreover, there is some \math{y'\in\posN} with \bigmaths{y' p\tightequal y}.
Thus \bigmaths{y'\tightprec y}.
Moreover, there is some \math{x'} with \bigmaths{x' p\tightequal x}.
Then \bigmaths{x'\nmid y'}.
It suffices to show that for any prime number \nlbmath q 
and any \math{n\tightin\posN} with \bigmaths{q^n\mid x'}{}
we have \bigmaths{q^n\mid y'}.
\\\underline{\math{q\tightequal p}:} \ 
From \bigmaths{p^n\mid x'}{} we get \bigmaths{p^{n+1}\mid x},
and then \bigmaths{p^{n+1}\mid y},
\ie\ \bigmaths{p^{n+1}\mid y' p}.
By \cororef{corollary trivial two}, we get \bigmaths{p^n\mid y'}.
\\\underline{\math{q\tightnotequal p}:} \ 
From \bigmaths{q^n\mid x'}{} we get \bigmaths{q^n\mid x},
and then \bigmaths{q^{n}\mid y},
\ie\ \bigmaths{q^{n}\mid y' p}.
By \lemmref{lemma new one}, we get \bigmaths{q^n\mid y'}.
\end{proofqed}

\yestop\begin{proofqed}{\lemmref{lemma Euclid VIII 14}} \ \ 
The ``only if''-direction is trivial. \
For the ``if''-direction, 
assume \bigmaths{x^2\mid y^2}{} and \bigmaths{p^n\mid x}{}
for arbitrary prime number \nlbmath p and \math{n\in\posN}. \ 
By \lemmref{lemma divides} it suffices to show \bigmaths{p^n\mid y}.
But from \bigmaths{p^n\mid x}{} we get \bigmaths{p^{2n}\mid x^2},
and then \bigmaths{p^{2n}\mid y^2}.
By \lemmref{lemma new one}, we get \bigmaths{p^n\mid y}.
\end{proofqed}

\yestop\begin{proofqed}{\lemmref{lemma additional one}} \ \ 
If any of \math{a,b} is equal to \nlbmath 0, then the other divides both and
thus (as \math{a,b} are coprime) must be equal to \nlbmath 1; 
and \bigmaths{x\tightequal 0};
in which case the lemma follows 
by \bigmaths{y:=a}{} and \bigmaths{z:=b}. 
If none of \math{a,b} is equal to \nlbmath 0, then also \math x is not 
equal to \nlbmath 0 and \math{a,x,b} are in continued proportion; \hskip.3em
so the lemma follows from 
\lemmref{lemma Euclid VIII 2} by setting its \nlbmath{n:=1}.\end{proofqed}

\yestop\begin{proofqed}{\lemmref{lemma coprime}} \ \ 
The ``if''-direction is trivial. For the ``only if''-direction let us assume
that \bigmaths{l_1,\ldots,l_n}{} are not coprime. Then there is some 
\nlbmath{x\not=1}
such that \bigmaths{x\mid l_i}{} for all \math{i\in\{1,\ldots,n\}}.
By \lemmref{lemma Euclid VII 31} there is some prime number \nlbmath p  
with \bigmaths{p\mid x}. By transitivity of \nlbmath\mid\ according to
\cororef{corollary divides}, we get \bigmaths{p\mid l_i}{} for all 
\math{i\in\{1,\ldots,n\}}.
\end{proofqed}

\yestop\begin{proofqed}{\lemmref{lemma coprime two}}
\initial{\underline{(1):}}
Suppose neither 
\math{p a,b} nor \math{a, p b} coprime.
By \lemmref{lemma coprime} there are two prime numbers \math{x,y}
with 
\bigmaths{x\mid p a},
\bigmaths{x\mid b},
\bigmaths{y\mid a},
\bigmaths{y\mid p b}.
As \math{a,b} coprime, \bigmaths{x\nmid a}.
As \math x is prime,
by \lemmref{lemma Euclid VII 30} we get \bigmaths{x\mid p}{} and thus 
\bigmaths{x\tightequal p}, as \math p is prime.
Similarly we get \bigmaths{y\tightequal p}. Thus \bigmaths{p\mid b}{} and
\bigmaths{p\mid a}, contradicting being \math{a,b} being coprime.
\initial{\underline{(2):}}
If \math{pa,b} are coprime,
they must be squares by
\lemmref{lemma additional one}, say \bigmaths{pa\tightequal l^2}{}
and \ \mbox{\math{b\tightequal k^2}}. \  
By \lemmref{lemma Euclid VII 30}, 
there is some \math m with \ \mbox{\math{p m\tightequal l}}, 
\ie\ \bigmaths{a\tightequal p m^2}.
By \lemmref{lemma Euclid VII 23}, \ 
\math{p m,k} \nolinebreak are coprime.
Thus, \bigmaths{k\tightin\posN}{} because \bigmaths{p m\tightnotequal 1}.
Moreover,
\bigmaths{\{p m^2,k^2\}=\{a,b\}}.
Similarly, 
if \math{a,p b} are coprime,
they must be squares by
\lemmref{lemma additional one}, say \bigmaths{a\tightequal k^2}{}
and \ \mbox{\math{p b\tightequal l^2}}. \  
By \lemmref{lemma Euclid VII 30}, there is some \math m with \ \mbox{\math
{p m\tightequal l}}, \ie\ \bigmaths{b\tightequal p m^2}.
Then again we have \bigmaths{k\tightin\posN},
\math{p m,k} are coprime, \ 
and \bigmaths{\{p m^2,k^2\}=\{a,b\}}.
\end{proofqed}

\yestop\begin{proofqed}{\lemmref{lemma mahoney footnote one two}}
\initial{\underline{(1):}}
By \cororef{corollary trivial one} we get
\bigmaths{x\mid\inpit{a\tight+b}\pm\inpit{a\tight-b}},
\ie\ \bigmaths{x\mid 2a}{} and 
\bigmaths{x\mid 2b}.
\initial{\underline{(2):}}
By (1) we have \bigmaths{x\mid 2a}{} and \bigmaths{x\mid 2b}. 
Assume \bigmaths{x\tightsucc 2}{} to show a contradiction.
If \bigmaths{2\mid x}, then there is some 
\math{k\tightsucceq 2} with \bigmaths{2k\tightequal x},
and then we have \bigmaths{k\mid a}{} and \bigmaths{k\mid b}, 
contradicting \math a, \math b being coprime. \ 
Otherwise, if \bigmaths{2\nmid x}, then by \lemmref{lemma Euclid VII 31}
there is some prime number \nlbmath{p\succ 2} with \bigmaths{p\mid x}.
Then \bigmaths{p\mid 2a}. By \lemmref{lemma Euclid VII 30} we get
\bigmaths{p\mid a}. Similarly \bigmaths{p\mid b}. This again contradicts 
\math a, \math b being coprime.
\end{proofqed}

\yestop\yestop\noindent 
The following lemma will be applied exclusively in the proof of
\lemmref{lemma p and q coprime}.
\begin{lemma}\label{lemma mahoney footnote one one} \ 
If \bigmaths{p\tightsucceq q}{}
and \bigmaths{x\mid 2p q}, then 
\bigmaths{x\mid p^2\tight-q^2}{} \uiff\ 
\bigmaths{x\mid p^2\tight+q^2}.\end{lemma}
\begin{proofqed}{\lemmref{lemma mahoney footnote one one}} \ \ 
As \bigmaths{x^2\mid 4p^2q^2}{} by \lemmref{lemma Euclid VIII 14}, 
the following are logically equivalent by
\cororef{corollary trivial one} and \lemmref{lemma Euclid VIII 14}:
\bigmaths{x\mid p^2\tight-q^2}; \ 
\bigmaths{x^2\mid\inpit{p^2\tight-q^2}^2}; \ 
\ \mbox{\maths{x^2\mid p^4\tight-2p^2q^2\tight+q^4};} \ \ 
\ \mbox{\maths{x^2\mid p^4\tight+2p^2q^2\tight+q^4};} \ \ 
\bigmaths{x^2\mid\inpit{p^2\tight+q^2}^2}; \ 
\bigmaths{x\mid p^2\tight+q^2}.
\end{proofqed}

\yestop\begin{proofqed}{\lemmref{lemma p and q coprime}} \ 
Suppose the contrary. Then 
by \lemmref{lemma coprime} there is a prime number \nlbmath x
with \bigmaths{x\mid p q}{} and \bigmaths{\inpit{x\mid p^2\tight+q^2}
\oder\inpit{x\mid p^2\tight-q^2}}.
Then we have \bigmaths{x\mid 2p q}, 
and then,
by \lemmref{lemma mahoney footnote one one},
we have \bigmaths{\inpit{x\mid p^2\tight+q^2}
\und\inpit{x\mid p^2\tight-q^2}}.
Moreover, by \lemmref{lemma mahoney footnote one two}(1),
we have
\bigmaths{x\mid 2p^2}{} and 
\bigmaths{x\mid 2q^2}.
As \math p, \math q are coprime, one of them is odd.
Thus, one of \math{p q} and \math{p^2\tight+q^2} is odd.
Thus, as \math x divides both, \bigmaths{x\tightnotequal 2}.
Thus, as \math x is prime, 
we have \bigmaths{x\nmid 2}. 
By \lemmref{lemma Euclid VII 30}, we get
\bigmaths{x\mid p^2}{} and 
\bigmaths{x\mid q^2}, and then \bigmaths{x\mid p}{} and 
\bigmaths{x\mid q}, contradicting \math p, \math q being coprime.
\end{proofqed}\vfill\pagebreak

\vfill\pagebreak
\yestop\yestop\halftop

\nocite{writing-mathematics}
\addcontentsline{toc}{section}{\refname}
\bibliography{herbrandbib}

\end{document}